%% file: main.tex
\documentclass[sigconf]{acmart}

\input{macros}

\pdfoutput=1

\copyrightyear{2017} 
\acmYear{2017} 
\setcopyright{acmlicensed}
\acmConference{CIKM'17 }{November 6--10, 2017}{Singapore, Singapore}
\acmPrice{15.00}
\acmDOI{10.1145/3132847.3132877}
\acmISBN{978-1-4503-4918-5/17/11}

\title{CSI:  A Hybrid Deep Model for Fake News Detection}

\author{Natali Ruchansky}
\authornote{These authors contributed equally to this work.}
\affiliation{%
  \institution{University of Southern California}
  \city{Los Angeles} 
  \state{California} 
}
\email{natalir@bu.edu}

\author{Sungyong Seo{\setcounter{footnote}{0}}}
\authornoteN{}
\affiliation{%
  \institution{University of Southern California}
  \city{Los Angeles} 
  \state{California} 
}
\email{sungyons@usc.edu}

\author{Yan Liu}
\affiliation{%
  \institution{University of Southern California}
  \city{Los Angeles} 
  \state{California} 
}
\email{yanliu.cs@usc.edu}

\begin{document}

\begin{abstract}

The topic of fake news has drawn attention both from the public and the academic communities. Such misinformation has the potential of affecting public opinion, providing an opportunity for malicious parties to manipulate the outcomes of public events such as elections.
Because such high stakes are at play,  
automatically detecting fake news is an important, yet challenging problem that is not yet well understood.
Nevertheless, there are three generally agreed upon characteristics of fake news: the \emph{text} of an article, the user \emph{response} it receives, and the \emph{source} users promoting it.   Existing work has largely focused on tailoring solutions to one particular characteristic which has limited their success and generality.

In this work, we propose a model that combines all three characteristics for a more accurate and automated prediction.  Specifically, we incorporate the behavior of both parties, users and articles, and the group behavior of users who propagate fake news.  Motivated by the three characteristics, we propose a model called {\ourmodel} which is composed of three modules: {\modone}, {\modtwo}, and {\modthree}.  
The first module is based on the \emph{response} and \emph{text}; it uses a Recurrent Neural Network to capture the temporal pattern of user activity on a given article. The second module learns the \emph{source} characteristic based on the behavior of users, and the two are integrated with the third module to classify an article as fake or not. Experimental analysis on real-world data demonstrates that {\ourmodel} achieves higher accuracy than existing models, and extracts meaningful latent representations of both users and articles.

\end{abstract}


\maketitle    

\section{Introduction}\label{sec:intro}
\input{sections/intro}

\section{Related work}\label{sec:related}
\input{sections/related}

\section{Problem}\label{sec:problem}
\input{sections/problem}

\section{Model}\label{sec:model}
\input{sections/model}

\section{Experiments}\label{sec:experiments}
\input{sections/experiments}

\section{Conclusion}\label{sec:conclustion}
\input{sections/conclusion}

\section*{Acknowledgments}
This work is supported in part by NSF Research Grant IIS-1619458 and IIS-1254206. The views and conclusions are those of the authors and should not be interpreted as representing
the official policies of the funding agency, or the U.S. Government.

\Urlmuskip=0mu plus 0mu\relax
\bibliographystyle{ACM-Reference-Format}
\bibliography{reference,natali}

\end{document}

%% file: macros.tex
\usepackage{booktabs} 
\usepackage{graphicx}
\usepackage{amsmath}
\usepackage{capt-of}
\usepackage{subcaption}
\usepackage{varwidth}
\usepackage{url}
\newsavebox\tmpbox

\newcommand{\nredit}[1]{{\color{black} #1}}

\newcommand{\spara}[1]{\smallskip\noindent{\bf #1}}

\newcommand{\etal}{{et al.}}

\newcommand{\ourmodel}{{\tt CSI}}
\newcommand{\ourmodelx}{{\tt CI}}
\newcommand{\ourmodelxm}{{\tt CI-t}}
\newcommand{\scorew}{suspiciousness}

\newcommand{\score}{{\scorew} score}
\newcommand{\userfeat}{\ensuremath{\bb{y}_i}}
\newcommand{\articlevec}{\ensuremath{\bb{v}_j}}
\newcommand{\uservec}{\ensuremath{\tilde{\bb{y}}_i}}
\newcommand{\userveci}[1]{\ensuremath{\tilde{\bb{y}}_{#1}}}

\newcommand{\userscore}{\ensuremath{s_i}}
\newcommand{\scoreart}{\ensuremath{p_j}}
\newcommand{\resultant}{\ensuremath{\bb{c}_j}}

\newcommand{\modone}{{\tt Capture}}
\newcommand{\modtwo}{{\tt Score}}
\newcommand{\modthree}{{\tt Integrate}}

\newcommand{\twitter}{{\sc Twitter}}

\newcommand{\weibo}{{\sc Weibo}}

\newcommand{\bb}[1]{\textbf#1}

\usepackage[multiple]{footmisc}

\DeclareCaptionLabelFormat{andtable}{#1~#2  \&  \tablename~\thetable}

%% file: sections/intro.tex

\begin{figure*}[t]
    \centering
    \includegraphics[width=0.8\textwidth]{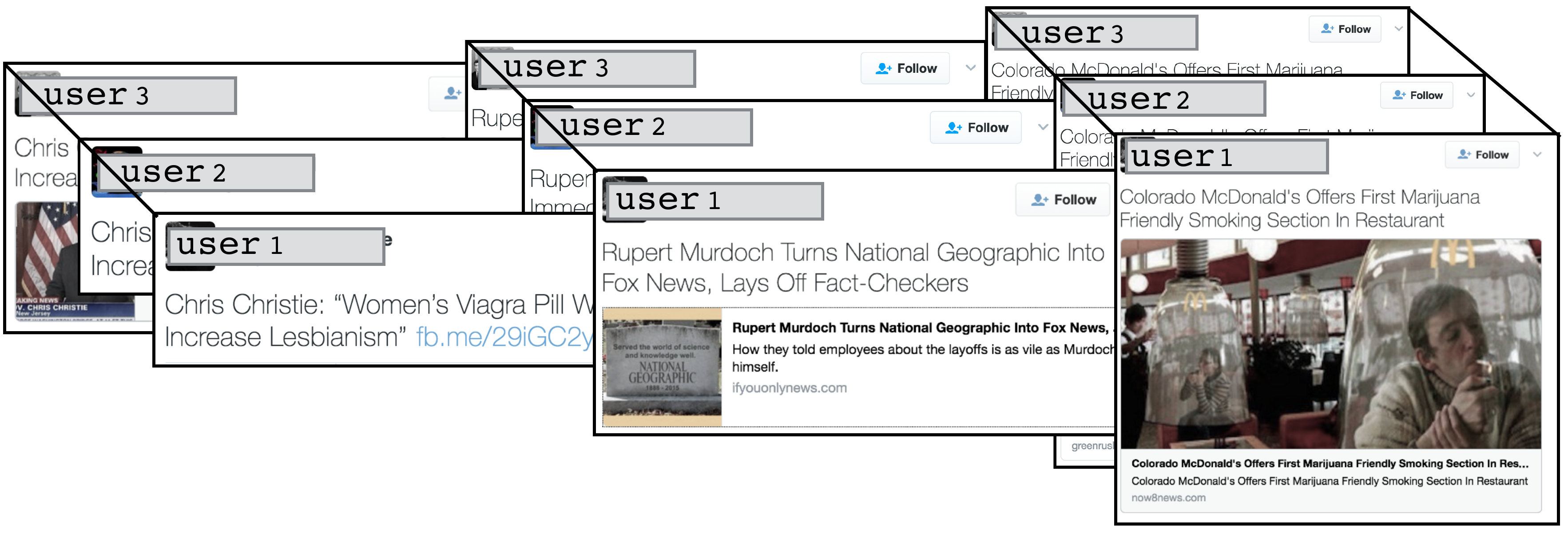}
    \caption{A group of Twitter accounts who shared the same set of fake articles.}
    \label{fig:fake-example}
\end{figure*}

Fake news on social media has experienced a resurgence of interest due to the recent political climate and the growing concern around its negative effect.  For example, in January 2017, a spokesman for the German government stated that they \emph{``are dealing with a phenomenon of a dimension that [they] have not seen before"}, referring to the proliferation of fake news~\cite{german}.    
Not only does it provide a source of spam in our lives, but fake news also has the potential to manipulate public perception and awareness in a major way.

Detecting misinformation on social media is an extremely important but also a technically challenging problem.  The difficulty comes in part from the fact that even the human eye cannot accurately distinguish true from false news; for example, one study found that  when shown a fake news article, respondents found it \emph{```somewhat' or `very' accurate 75\% of the time"}, and another found that 80\% of high school students had a hard time determining whether an article was fake~\cite{forbes,npr}.
 In an attempt to combat the growing misinformation and confusion, several \emph{fact-checking} websites have been deployed to expose or confirm stories (e.g. \texttt{snopes.com}).  These websites play a crucial role in combating fake news, but they require expert analysis which inhibits a timely response.  
As a response, numerous articles and blogs have been written to raise public awareness and provide tips on differentiating truth from falsehood~\cite{ted}.  
While each author provides a different set of signals to look out for, there are several characteristics that are generally agreed upon, relating to the \emph{text} of an article, the \emph{response} it receives, and its \emph{source}.

The most natural characteristic is the \emph{text} of an article.  Advice in the media varies from evaluating whether the headline matches the body of the article, to judging the consistency and quality of the language.  Attempts to automate the evaluation of \emph{text} have manifested in sophisticated natural language processing and machine learning techniques that rely on hand-crafted and data-specific textual features to classify a piece of text as true or false~\cite{gupta2014tweetcred,ferreira2016emergent,ma2016detecting,markowitz2014linguistic,markines2009social,rubin2015deception}.   
 These approaches are limited by the fact that
 the linguistic characteristics of fake news are still not yet fully understood.  Further, the characteristics vary across different types of fake news, topics, and media platforms.

A second characteristic is the \emph{response} that a news article is meant to illicit.  Advice columns encourage readers to consider how a story makes them feel -- does it provoke either anger or an emotional response? The advice stems from the observation that fake news often contains opinionated and inflammatory language, crafted as click bait or to incite confusion~\cite{chen2015misleading,rubin2017deception}.    
For example, the New York Times cited examples of people profiting from publishing fake stories online; the more provoking, the greater the response, and the larger the profit~\cite{nyt}. 
 Efforts to automate response detection typically model the spread of fake news as an epidemic on a social graph~\cite{friggeri2014rumor,jin2013epidemiological,starbird2014rumors,kumar2016disinformation}, or use hand-crafted features that are social-network dependent, such as the number of Facebook likes, combined with a traditional classifier~\cite{castillo2011information,ma2015detect,kwon2017rumor,wu2015false,zhao2015enquiring,markines2009social}.    Unfortunately, access to a social graph is not always feasible in practice, and manual selection of features is labor intensive.

A final characteristic is the \emph{source} of the article.  Advice here ranges from checking the structure of the url, to the credibility of the media source, to the profile of the journalist who authored it; in fact, Google has recently banned nearly 200 publishers to aid this task~\cite{google}.   
In the interest of exposure to a large audience, 
a set of loyal promoters may be deployed to publicize and disseminate the content.  In fact, several small-scale analyses have observed that there are often groups of users that heavily publicize fake news, particularly just after its publication~\cite{points,kremlin}.  
For example, Figure~\ref{fig:fake-example} shows an example of three Twitter users who consistently promote the same \emph{fake} news stories.  Approaches here typically focus on data-dependent user behaviors, or identifying the source of an epidemic, and disregard the fake news articles themselves~\cite{wang2014rumor,mukherjee2012spotting}.

Each of the three characteristics mentioned above has ambiguities that make it challenging to successfully automate fake news detection based on just one of them.  Linguistic characteristics are not fully understood, hand-crafted features are data-specific and arduous, and source identification does not trivially lead to fake news detection.
In this work, we build a more accurate automated fake news detection by utilizing all three characteristics at once: \emph{text}, \emph{response}, and \emph{source}.
Instead of relying on manual feature selection, the {\ourmodel} model that we propose is built upon deep neural networks, which can automatically select important features.  Neural networks also enable {\ourmodel} to exploit information from different domains and capture temporal dependencies in users engagement with articles.
A key property of {\ourmodel} is that it explicitly outputs information both on articles \emph{and} users, and does not require the existence of a social graph, domain knowledge, nor assumptions on the types and distribution of behaviors that occur in the data.

Specifically,  {\ourmodel} is composed of one module for each side of the activity, user \emph{and} article -- Figure~\ref{fig:model_intuition} illustrates the intuition.  The first module, called {\modone},  exploits the temporal pattern of user activity, including text, to capture the \emph{response} a given article received.  {\modone} is constructed as a Recurrent Neural Network (more precisely an LSTM) which receives article-specific information such as the temporal spacing of user activity on the article and a doc2vec~\cite{le2014distributed} representation of the text generated in this activity (such as a tweet).
The second module, which we call {\modtwo}, uses a neural network and an \nredit{implicit user graph to extract a representation and assign a score} to each user that is indicative of their propensity to participate in a \emph{source} promotion group.  Finally, the third module, {\modthree}, combines the \emph{response}, \emph{text}, and \emph{source} information from the first two modules to classify each article as fake or not.
The three module composition of {\ourmodel} allows it to independently learn characteristics from both sides of the activity, combine them for a more accurate prediction and output feedback both on the articles (as a falsehood classification) and on the users (as a suspiciousness score).

Experiments on two real-world datasets demonstrate that by incorporating \emph{text}, \emph{response}, and \emph{source}, the  {\ourmodel} model achieves significantly higher classification accuracy than existing models.  In addition, we demonstrate that both the {\modone} and {\modtwo} modules provide meaningful information on each side of the activity.  {\modone} generates low-dimensional representations of news articles and users that can be used for tasks other than classification, and {\modtwo} rates users by their participation in group behavior.  

\noindent
The main contributions can be summarized as:
\begin{enumerate}
	\item To the best of our knowledge, we propose the first model that explicitly captures the three common characteristics of fake news, \emph{text}, \emph{response}, and \emph{source},  and identifies misinformation both on the article and on the user side.
	\item The proposed model, which we call {\ourmodel}, evades the cost of manual feature selection by incorporating neural networks.  The features we use capture the temporal behavior and textual content in a general way that does not depend on the data context nor require distributional assumptions.
	\item Experiments on real world datasets demonstrate that {\ourmodel} is more accurate in fake news classification than previous work, while requiring fewer parameters and training.
\end{enumerate}

\vspace{-.18cm}

%% file: sections/related.tex

\nredit{
The task of detecting fake news has undergone a variety of labels, from misinformation, to rumor, to spam.   
Just as each individual may have their own intuitive definition of such related concepts, each paper adopts its own definition of these words which conflicts or overlaps both with other terms and other papers.  For this reason, we specify that the target of our study is detecting news content that is fabricated, that is fake.  Given the disparity in terminology, we overview existing work grouped loosely according to which of the three characteristics (\emph{text}, \emph{response}, and \emph{source}) it considers.}

There has been a large body of work surrounding \emph{text} analysis of fake news and similar topics such as rumors or spam.  This work has focused on mining particular linguistic cues, for example, by finding anomalous patterns of pronouns, conjunctions, and words associated with negative emotional word usage~\cite{feng2013detecting, markowitz2014linguistic}.  For example, Gupta {\etal}~\cite{gupta2014tweetcred} found that fake news often contain an inflated number of swear words and personal pronouns.    Branching off of the core linguistic analysis, many have combined the approach with traditional classifiers to label an article as true or false~\cite{castillo2011information,ma2015detect,kwon2017rumor,wu2015false,zhao2015enquiring,markines2009social,ferreira2016emergent}.
Unfortunately, the linguistic indicators of fake news across topic and media platform are not yet well understood; Rubin {\etal}~\cite{rubin2015deception}  explained that there are many types of fake news, each with different potential textual indicators.   Thus existing works design hand-crafted features which is not only laborious but highly dependent on the specific dataset and the availability of domain knowledge to design appropriate features.  To expand beyond the specificity of hand-crafted features, Ma {\etal}~\cite{ma2016detecting} proposed a model based on recurrent neural networks that uses mainly linguistic features.  In contrast to~\cite{ma2016detecting}, the {\ourmodel} model we propose captures all three characteristics, is able to isolate suspicious users, and requires fewer parameters for a more accurate classification.

The \emph{response} characteristic has also received attention in existing work.  Outside of the fake news domain, Castillo {\etal}~\cite{castillo2014characterizing} showed that the temporal pattern of user response to news articles plays an important role in understanding the properties of the content itself.  From a slightly different point of view, one popular approach has been to measure the response an article received by studying its propagation on a social graph~\cite{friggeri2014rumor,jin2013epidemiological,starbird2014rumors,kumar2016disinformation}.  The epidemic approach requires access to a graph which is infeasible in many scenarios.  Another approach has been to utilize hand-crafted social-network dependent behaviors, such as the number of Facebook likes, as features in a classifier~\cite{castillo2011information,ma2015detect,kwon2017rumor,wu2015false,zhao2015enquiring,markines2009social}.  
  As with the linguistic features, these works require feature-engineering which is laborious and lacks generality.

The final characteristic, \emph{source}, has been studied as the task of identifying the source of an epidemic on a graph~\cite{wang2014rumor,luo2013identifying,zhu2016information}, or isolating bots based on certain documented behaviors~\cite{chavoshidebot,varol2017online}.  Another approach identifies group anomalies.  Early work in group anomaly detection assumed that the groups were known a priori, and the goal was to detect which of them were anomalous~\cite{mukherjee2012spotting}. Such information is not feasible in practice, hence later works propose variants of mixtures models for the data, where the learned parameters are used to identify the anomalous groups~\cite{xiong2011group,xiong2011hierarchical}.  Muandet \etal{}~\cite{muandet2013one} took a similar approach by combining kernel embedding with an SVM classifier.  Most recently, Yu {\etal}~\cite{yu2015glad} proposed a unified hierarchical Bayes model to infer the groups and detect group anomalies simultaneously. \nredit{There has also been a strong line of work surrounding detecting suspicious user behavior of various types; a nice overview is given in~\cite{jiang2016suspicious}.  Of this line, the most related is the CopyCatch model proposed in~\cite{beutel2013copycatch}, which identifies temporal bipartite cores of user activity on pages.} In contrast to existing works, the {\ourmodel} model we propose can identify group anomalies as well as the core behaviors they are responsible for (fake news). The model does not require group information as input, does not make assumptions about a particular distribution, \nredit{and learns a representation and score for each user.}

In contrast to the vast array of work highlighted here, the {\ourmodel} model we propose does not rely on hand-crafted features, domain knowledge, or distributional assumptions, offering a more general modeling of the data.  Further, {\ourmodel} captures all three characteristics and outputs both a classification of articles, a scoring of users, and representations of both users and articles that can be used for in separate analysis.

%% file: sections/problem.tex

In this section we first lay out preliminaries, and then discuss the context of fake news which we address.

\begin{figure}[h]
    \centering
    \includegraphics[width=0.3\textwidth]{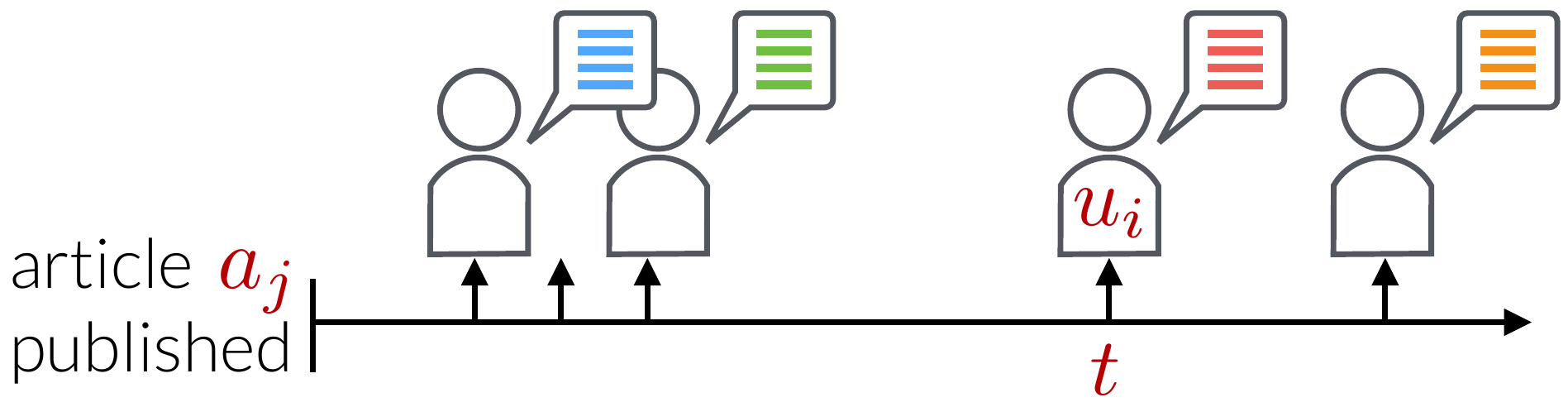}
    \caption{Temporal engagements of users with articles.}
    \label{fig:dadta}
\end{figure}

\spara{Preliminaries:}   We consider a series of temporal \emph{engagements} that occurred between $n$ users with $m$ news-articles over time $[1,T]$.  Each engagement between a user $u_i$ and an article $a_j$ at time $t$ is represented as $e_{ijt}=(u_i, a_j,t)$.  In particular, in our setting, an engagement is composed of textual information relayed by the user $u_i$ about article $a_j$, at time $t$; for example, a tweet or a Facebook post.  Figure~\ref{fig:dadta} illustrates the setting.  In addition, we assume that each news article is associated with a label $L(a_j)=0$ if the news is true, and $L(a_j)=1$ if it is false.
Throughout we will use italic characters $x$ for scalars, bold characters $\bb{h}$ for vectors, and capital bold characters $\bb{W}$ for matrices. 

\spara{Goal:}  While the overarching theme of this work is fake news detection, the goal is two fold (1) accurately classify fake news, and (2) identify groups of suspicious users.
In particular, given a temporal sequence of engagements $E=\{e_{ijt}=(u_i,a_j,t)\}$, our goal is to  produce a label $\hat{L}(a_j)\in[0,1]$ for each article, and a suspiciousness score $s_i$ for each user.  To do this we encapsulate the \emph{text}, \emph{response}, and \emph{source} characteristics in a model and capture the temporal behavior of both parties, users and articles, as well as textual information exchanged in the activity.
We make no assumptions on the distribution of user behavior, nor on the context of the engagement activity. 


%% file: sections/model.tex

\begin{figure*}[h]
  \captionsetup[subfigure]{justification=centering}
  \begin{subfigure}{0.47\textwidth}
  \centering
    \includegraphics[width=0.9\textwidth]{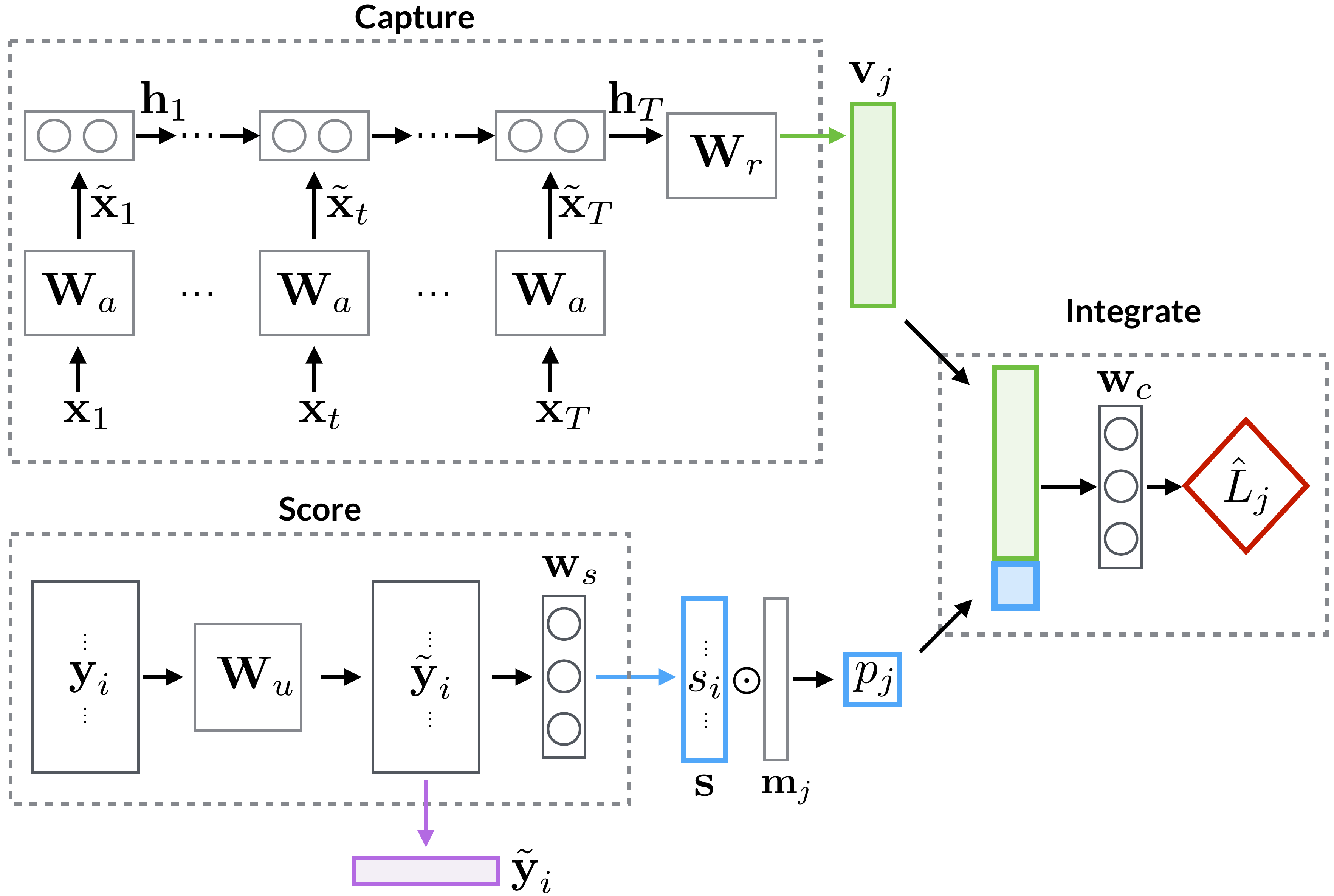}
\caption{The {\ourmodel} model specification.  The {\modone} module depicts the LSTM for a single article $a_j$, while the {\modtwo} module operates over all users.  The output of {\modtwo} is then filtered to be relevant to $a_j$.}
\end{subfigure}
\hspace*{\fill} 
\begin{subfigure}{0.47\textwidth}
  \centering
        \includegraphics[width=0.85\textwidth]{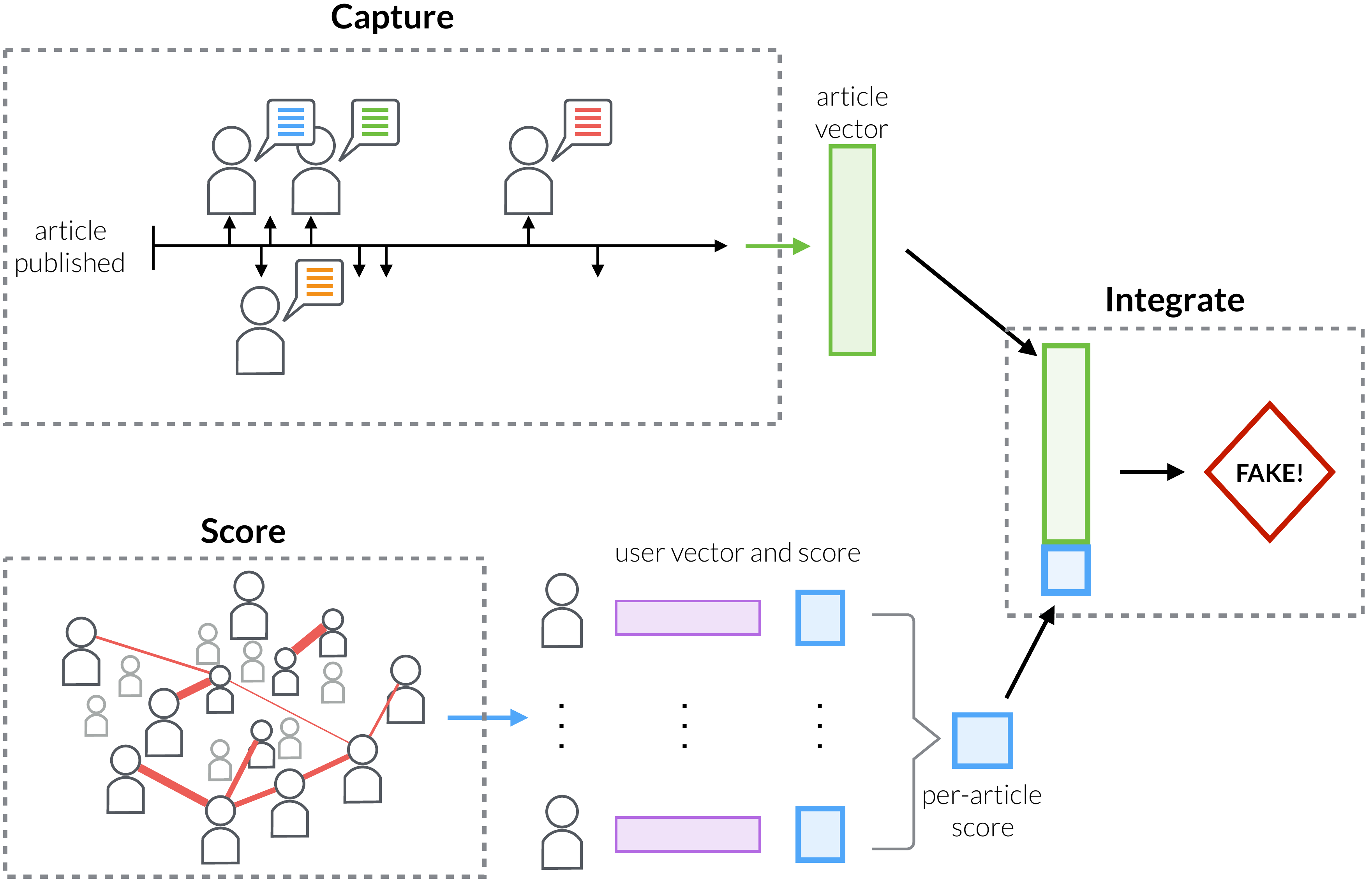}\vspace{.6cm}
\caption{Intuition behind {\ourmodel}.  Here, {\modone} receives the temporal series of engagements, and {\modtwo} is fed an implicit user graph constructed from the engagements over all articles in the data.\label{fig:model_intuition}}
\end{subfigure}
  \caption{An illustration of the proposed {\ourmodel} model. \label{fig:model}}
  \end{figure*}

In this section, we give the details of the proposed model, which we call {\ourmodel}. The model consists of two main parts, a module for extracting temporal representation of news articles, \nredit{and a module for representing and scoring the behavior of users.} 
The former captures the \emph{response} characteristic described in Section~\ref{sec:intro} while incorporating \emph{text}, and the latter captures the \emph{source} characteristic.  Specifically, {\ourmodel} is composed of the following three parts, \nredit{the specification and intuition of which is shown} in Figure~\ref{fig:model}:
\begin{enumerate}
	\item {\modone}: To extract temporal representations of articles we use a Recurrent Neural Network (RNN). Temporal engagements are stored as vectors and are fed into the RNN which produces an output a representation vector $\articlevec$. 
	\item {\modtwo}:  \nredit{To compute a score $\userscore $ and representation $\uservec$, user-features are fed into a fully connected layer and a weight is applied to produce the scores vectors \bb{s}. }
	\item {\modthree}: The outputs of the two modules are concatenated and the resultant vector is used for classification.

\end{enumerate}

\noindent
\nredit{With the first two modules, {\modone} and {\modtwo}, the {\ourmodel} model extracts representations of both users and articles as low-dimensional vectors; these representations are important for the fake news task, but can also be used for independent analysis of users and articles.  In addition, {\modtwo} produces a score for each user as a compact version of the vector.  The {\modthree} module then combines the article representations with the user scores for an ultimate prediction of the veracity of an article.} 
In the sections that follow, we discuss the details of each module.

\subsection{{\modone} news article representation}
In the first module, we seek to capture the pattern of temporal engagement of users with an article $a_j$ both in terms of the frequency and distribution.  \nredit{In other words, we wish to capture not only the number of users that engaged with $a_j$ in Figure~\ref{fig:model_intuition}, but also how the engagements were spaced over time.}  Further, we incorporate textual information naturally available with the engagement, such as the text of a tweet, in a general and automated way.

As the core of the first module, we use a Recurrent Neural Network (RNN), since RNNs
 have been shown to be effective at capturing temporal patterns in data and for integrating different sources of information.  A key component of {\modone} is the choice of features used as input to the cells for each article.   Our feature vector $\bb{x}_t$ has the following form:
  \[ \bb{x}_t=(\eta,\Delta t,\bb{x}_u,\bb{x}_\tau)\]
The first two variables, $\eta$ and $\Delta t$,  capture the temporal pattern of engagement an article receives with two simple, yet powerful quantities:  the number of engagements $\eta$, and the time between engagements $\Delta t$.  
 Together, $\eta$ and $\Delta t$ provide a general of measure the frequency and distribution of the \emph{response} an article received.  
 Next, we incorporate \emph{source} by adding a user feature vector $\bb{x}_u$ that is global and not specific to a given article.  In line with existing literature on  information retrieval and recommender systems~\cite{leskovec2014mining}, we construct the binary incidence matrix of which articles a user engaged with, and apply the Singular Value Decomposition (SVD) to extract a lower-dimensional representation for each $u_i$. 
 Finally, a vector $\bb{x}_\tau$ is included which carries the \emph{text} characteristic of an engagement with a given article $a_j$.  To avoid hand-crafted textual feature selection for $\bb{x}_\tau$, we use~\textit{doc2vec}~\citep{le2014distributed} on the text of each engagement.  Further technical details will be explained in Section~\ref{sec:experiments}.

Since the temporal and textual features come from different domains, it is not desirable to incorporate them into the RNN as raw input.
To standardize the input features, we insert an embedding layer between the raw features $\textbf{x}_t$ and the inputs $\tilde{\textbf{x}}_t$ of the RNN. This embedding layer is a fully connected layer as following:
\begin{align*}
\tilde{\textbf{x}}_t = \text{tanh}(\textbf{W}_a\textbf{x}_t+\textbf{b}_a)
\end{align*}
where $\textbf{W}_a$ is a weight matrix applied to the raw features $\textbf{x}_t$ at time $t$ and $\textbf{b}_a$ is a bias vector.  Both $\textbf{W}_a$  and $\textbf{b}_a$ are the fixed for all $\textbf{x}_t$.
\nredit{To capture the temporal \emph{response} of users to an article, we construct the {\modone} module using a Long Short-Term Memory (LSTM) model because of its propensity for capturing long-term dependencies and its flexibility in processing inputs of variable lengths.   }
For the sake of brevity we do not discuss the well-established LSTM model here, but refer the interested reader to~\citep{husken2003recurrent} for more detail.

What is important for our discussion is that in the final step of the LSTM, $\tilde{\textbf{x}}_T$ is fed as input and the last hidden state $\textbf{h}_T$ is passed to the fully connected layer.  The result is a vector: 
\[\articlevec= \text{tanh}(\textbf{W}_r\textbf{h}_T+\textbf{b}_r)\]
This vector serves as a low dimension representation of the temporal pattern of engagements a given article $a_j$ received\nredit{ -- capturing both the \emph{response} and \emph{textual} characteristics.  The vectors $\articlevec$ will be fed to the {\modthree} module for article classification, but can also be used for stand-alone analysis of articles.}

\spara{Partitioning: }
In principle, the feature vector $\bb{x}_t$ associated with each engagement can be considered as an input into a cell; however, this would be highly inefficient for large data. 
A more efficient approach is to partition a given sequence by changing the granularity, and using an aggregate of each partition (such as an average) as input to a cell. 
Specifically, the feature vector for article $a_j$ at partition $t$ has the following form:
$\eta$ is the number of engagements that occurred in partition $t$, $\Delta t$ holds the time between the current and previous non-empty partitions, $\bb{x}_u$ is the average of user-features over users $u_i$ that engaged with $a_j$ during $t$, and $\tau$ is the textual content exchanged during $t$.    

\subsection{{\modtwo} users}

\nredit{
In the second module, we wish to capture the \emph{source} characteristic present in the behavior of users.  To do this, we seek a compact representation that will have the same (small) dimension for every article (since it will ultimately be used in the {\modthree} module).   Given a set of user features, we first apply a fully connected layer to extract vector representations of each user as follows:
\[\uservec= \text{tanh}(\textbf{W}_u{\userfeat}+\textbf{b}_u)\]
where $\bb{W}_u$ is the weight matrix and $\bb{b}_u$ is the bias; $L2$-regularization is used on $\bb{W}_u$ with parameter $\lambda$. This results in a vector representation $\uservec$ for each user $u_i$ that is learned jointly with the {\modone} module.  To aggregate this information,  we apply a weight vector $\bb{w}_s$  to produce a scalar score $\userscore $ for each user as:
\[ \userscore =\sigma(\textbf{w}^\top_s\cdot \uservec  + {b}_s) \]
with ${b}_s$ as the bias of a fully connected layer, and  $\sigma$ as the sigmoid function.  The set of $\userscore $ forms the vector $\bb{s}$ of user scores.

In principle, user features can be constructed using information from the users social network profile.
Since we wish to capture the \emph{source} characteristic, we construct a weighted user graph where an edge denotes the number of articles with which two users have both engaged.  Users who engage in group behavior will correspond to dense blocks in the adjacency matrix.  Following the literature, we apply the SVD to the adjacency matrix and extract a lower-dimensional feature $\userfeat$  for each user, ultimately obtaining $(\userscore ,\uservec)$ for each user $u_i$.

By constructing the {\modtwo} module in this way, {\ourmodel} is able to jointly learn from the two sides of the engagements while extracting information that is meaningful to the \emph{source} characteristic.  As with the {\modone} module, the vector $\uservec$ can be used for stand-alone analysis of the users. }

\subsection{{\modthree} to classify}~\label{sec:integrate}

Each of the {\modone} and {\modtwo} modules outputs information on articles and users with respect to the three characteristics of interest.  In order to incorporate the two sources of information, we propose a third module as the final step of {\ourmodel} in which article representations $\articlevec$ are combined with the user scores $\userscore $ to produce a label prediction $\hat{L}_j$ for each article.  

To integrate the two modules, we apply a mask $\bb{m}_j$ to the vector $\bb{s}$ that selects only the entries $\userscore $ whose corresponding user $u_i$ engaged with a given article $a_j$.  These values are average to produce $\scoreart$ which captures the {\score} of the users that engage with the specific article $a_j$.
The overall score $\scoreart$ is concatenated with $\articlevec$ from {\modone}, and the resultant vector $\resultant$ is fed into the last fully connected layer to predict the label $\hat{L}_j$ of article $a_j$.
\begin{align*}
\hat{L}_j = \sigma(\textbf{w}^\top_c\resultant + {b}_c)
\end{align*}
\nredit{
This integration step enables the modules to work together to form a more accurate prediction.  By jointly training the {\ourmodel} with the {\modone} and {\modtwo} modules, the model learns both user and article information simultaneously.
At the same time, the {\ourmodel} model generates information on articles and users that captures different important characteristics of the fake news problem, and combines the information for an ultimate prediction.}

\spara{Training}:
The loss function for training {\ourmodel} is specified as:
\begin{align*}
Loss = -\frac{1}{N}\sum_{j=1}^{N}\left[L_j\log\hat{L}_J + (1-L_j)\log(1-\hat{L}_j)\right] + \frac{\lambda}{2}||\bb{W}_u||^2_2
\end{align*}
where $L_j$ is a the ground-truth label. To reduce overfitting in {\ourmodel}, random units in $\bb{W}_a$ and $\bb{W}_r$ are dropped out for training. Under these constraints, the parameters in {\modone}, {\modtwo}, and {\modthree} are jointly trained by back-propagation. 

\subsection{Generality}
 
 We have presented the {\ourmodel} model in the context of fake news; however, our model can be easily generalized to any dataset.
 Consider a set of engagements between an actor $q_i$ and a target $r_j$ over time $t\in[0,T]$, \nredit{ in other words, the article in Figure~\ref{fig:model_intuition} is a target and each user is an actor.}  The {\modone} module can be used to capture the temporal patterns of engagements exhibited on targets by actors, and {\modtwo} can be used to \nredit{extract a score and representation of} each actor $q_i$ that captures  the participation in group behavior.
 Finally, {\modthree} combines the first two modules to enhance the prediction quality on targets.
 For example, consider users accessing a set of databases.  The {\modone} module can identify databases which  received an unusual pattern of access, and {\modtwo} can highlight users that were likely responsible.
 In addition, the flexibility of {\ourmodel} allows for integration of additional domain knowledge.

%% file: sections/experiments.tex

\begin{table}[t]
\centering
\begin{tabular}{r| rr}
\toprule
          \multicolumn{1}{r}{}      &  \multicolumn{1}{r}{{\twitter}} & \multicolumn{1}{r}{{\weibo}} \\ \midrule
\# Users       &  233,719   &   2,819,338             \\ 
\# Articles      &  992    &     4,664    \\ 
\# Engagements &  592,391    &  3,752,459                   \\ 
\# Fake articles      & 498     &   2,313               \\ 
\# True  articles  & 494     &   2,351                    \\ 
Avg $T$ per article (hours) & 1,983  &1,808  \\           \bottomrule

\end{tabular}
\caption{Statistics of the datasets.\label{tab:stat}}
\end{table}

In this section, we demonstrate the quality of {\ourmodel} on two real world datasets. In the main set of experiments, we evaluate the accuracy of the classification produced by {\ourmodel}. In addition, we investigate the quality of the scores and representations produced by the {\modtwo} module and show that they are highly related to the \emph{score} characteristic. Finally, we show the robustness of our model when labeled data is limited and investigate temporal behaviors of suspicious users.

\spara{Datasets}
In order to have a fair comparison, we use two real-world social media datasets that have been used in previous work, {\sc {\twitter}} and {\weibo}~\cite{ma2016detecting}.  \nredit{To date, these are the only publicly available datasets that include all three characteristics: \emph{response, text}, and \emph{user} information.}
  Each dataset has a number of articles with labels $L(a_j)$; in {\twitter} the articles are news stories, and in {\weibo} they are discussion topics. Each article also has a set of engagements (tweets) made by a user $u_i$ at time $t$.
A summary of the statistics is listed in Table \ref{tab:stat}.


\subsection{Model setup}
We first describe the details of two important components in {\ourmodel}: 1) how to obtain the temporal partitions discussed in Section~\ref{sec:model} and 2) the specific features for each dataset.

\spara{Partitioning:} As mentioned in Section~\ref{sec:model}, treating each time-stamp as its own input to a cell can be extremely inefficient and can reduce utility.  Hence, we propose to partition the data into segments, each of which will be an input to a cell.  We apply a natural partitioning by changing the temporal granularity from \textit{seconds} to \textit{hours}. 

 \spara{Hyperparameters:}
We use cross-validation to set the regularization parameter for the loss function in Section~\ref{sec:integrate} to $\lambda=0.01$, the dropout probability as $0.2$, the learning rate to $0.001$, and use the Adam optimizer.

 \spara{Features:} 
Recall from Section~\ref{sec:model} that {\modone} operates on $\bb{x}_t=(\eta,\Delta t,\bb{x}_u,\bb{x}_\tau)$ -- temporal, user, and textual features.
To apply \textit{doc2vec}\cite{le2014distributed} to the {{\weibo}} data, we first apply Chinese text segmentation.\footnote{\url{https://github.com/fxsjy/jieba}}
To extract $\bb{x}_u$, we apply the SVD with rank $20$ for {\twitter} and $10$ for {\weibo}, resulting in $122$ dimensional $\bb{x}_t$ for {\twitter} and $112$ for {\weibo}. 
(SVD dimension chosen using the Scree plot.)
We then set the embedding dimension so that each $\tilde{\bb{x}}_t$ has dimension $100$.
The SVD rank for $\bb{x}_i$ for {\modtwo} is  $50$ for both datasets\nredit{, and the dimension of $\bb{W}_u$ is 100.}

\begin{table}[t]
\centering
\begin{tabular}{l c c c c}
\toprule
        & \multicolumn{2}{c}{{\twitter}}      & \multicolumn{2}{c}{{\weibo}}      \\ 
        \cmidrule(lr){2-3}   \cmidrule(lr){4-5}
   & Accuracy       & F-score       & Accuracy       & F-score       \\ \midrule
\textsc{DT-Rank} & 0.624          & 0.636          & 0.732          & 0.726          \\
\textsc{DTC}     & 0.711          & 0.702          & 0.831          & 0.831          \\
\textsc{SVM-TS}     & 0.767       & 0.773          & 0.857          & 0.861          \\
\textsc{LSTM-1}  & 0.814          & 0.808          & 0.896               & 0.913               \\
\textsc{GRU-2}   & 0.835    &   0.830         & 0.910          & 0.914          \\ \midrule
\ourmodelx    & 0.847          & 0.846          & 0.928          & 0.927 \\ 
\ourmodelxm    & 0.854          & 0.848          & 0.939          & 0.940 \\ 
\ourmodel    & \textbf{0.892} & \textbf{0.894} & \textbf{0.953} & \textbf{0.954} \\ \bottomrule
\end{tabular}
\caption{Comparison of detection accuracy on two datasets}
\label{tab:acc}
\end{table}

\subsection{Fake news classification accuracy}\label{sec:fake_acc}
In the main set of experiments, we use two real-world datasets, {\twitter} and {\weibo}, to compare the proposed {\ourmodel} model with five state-of-the-art  models that have been used for similar classification tasks and were discussed in Section~\ref{sec:related}:  \textsc{SVM-TS}~\citep{ma2015detect} , \textsc{DT-Rank}~\citep{zhao2015enquiring},  \textsc{DTC}~\cite{castillo2011information} ,  \textsc{LSTM-1}~\cite{ma2016detecting}, and  \textsc{GRU-2}~\cite{ma2016detecting}.
Further, to evaluate the utility of different features included in the model, we consider {\ourmodelx} as the {\ourmodel} model using only textual features $\bb{x}_t=(\bb{x}_\tau)$, {\ourmodelxm} as using textual and temporal features $\bb{x}_t=(\eta,\Delta t,\bb{x}_\tau)$, and finally {\ourmodel} using textual, temporal, and user features.   Since the first two do not incorporate user information, we omit the \textsc{S} from the name.
All RNN-based models including \textsc{LSTM-1} and \textsc{GRU-2} were implemented with Theano\footnote{\url{http://deeplearning.net/software/theano}} and tested with Nvidia Tesla K40c GPU. The AdaGrad algorithm is used as an optimizer for \textsc{LSTM-1} and \textsc{GRU-2} as per~\citep{ma2016detecting}. For {\ourmodel}, we used the Adam algorithm.

Table~\ref{tab:acc} shows the classification results using 80\% of entire data as training samples, 5\% to tune parameters, and the remaining 15\% for testing; we use 5-fold cross validation.  This division is chosen following previous work for fair comparison, and will be studied in later sections.  We see that {\ourmodel} outperforms other models in both accuracy and F-score. Specifically, {\ourmodelx} shows similar performance with \textsc{GRU-2} which is a more complex 2-layer stacked network. This performance validates our choice of capturing fundamental temporal behavior, and demonstrates how a simpler structure can benefit from better features and partitioning.
Further, it shows the benefit of utilizing \textit{doc2vec} over simple \textit{tf-idf}.

Next, we see that {\ourmodelxm} exhibits an improvement of more than $1\%$ in both accuracy and F-score over {\ourmodelx}.  This demonstrated that while linguistic features may carry some temporal properties, the  frequency and distribution of engagements caries useful information in capturing the difference between true and fake news.

Finally, {\ourmodel} gives the best performance over all comparison models and versions.
We see that integrating user features boosts the overall numbers up to 4.3\% from \textsc{GRU-2}.  Put together, these results demonstrate that {\ourmodel} successfully captures and leverages all three characteristics of \emph{text}, \emph{response}, and \emph{source}, for accurately classifying fake news.


\subsection{Model complexity}

In practice, the availability of labeled examples of true and fake news may be limited, hence, in this section, we study the usability of {\ourmodel} in terms of the number of parameters and amount of labeled training samples it requires.

Although {\ourmodel} is based on deep neural networks, the compact set of features that {\modone} utilizes results in fewer required parameters than other models. Furthermore, the user relations in \modtwo{} can deliver condensed representations which cannot be captured by an RNN, allowing \ourmodel{} to have less parameters than other RNN-based models.   In particular, the model has on the order of $52K$ parameters, whereas \textsc{GRU-2} has $621K$ parameter. 

\begin{figure}
  \captionsetup[subfigure]{justification=centering}
\begin{subfigure}{0.23\textwidth}
  \centering
    \includegraphics[scale=.19]{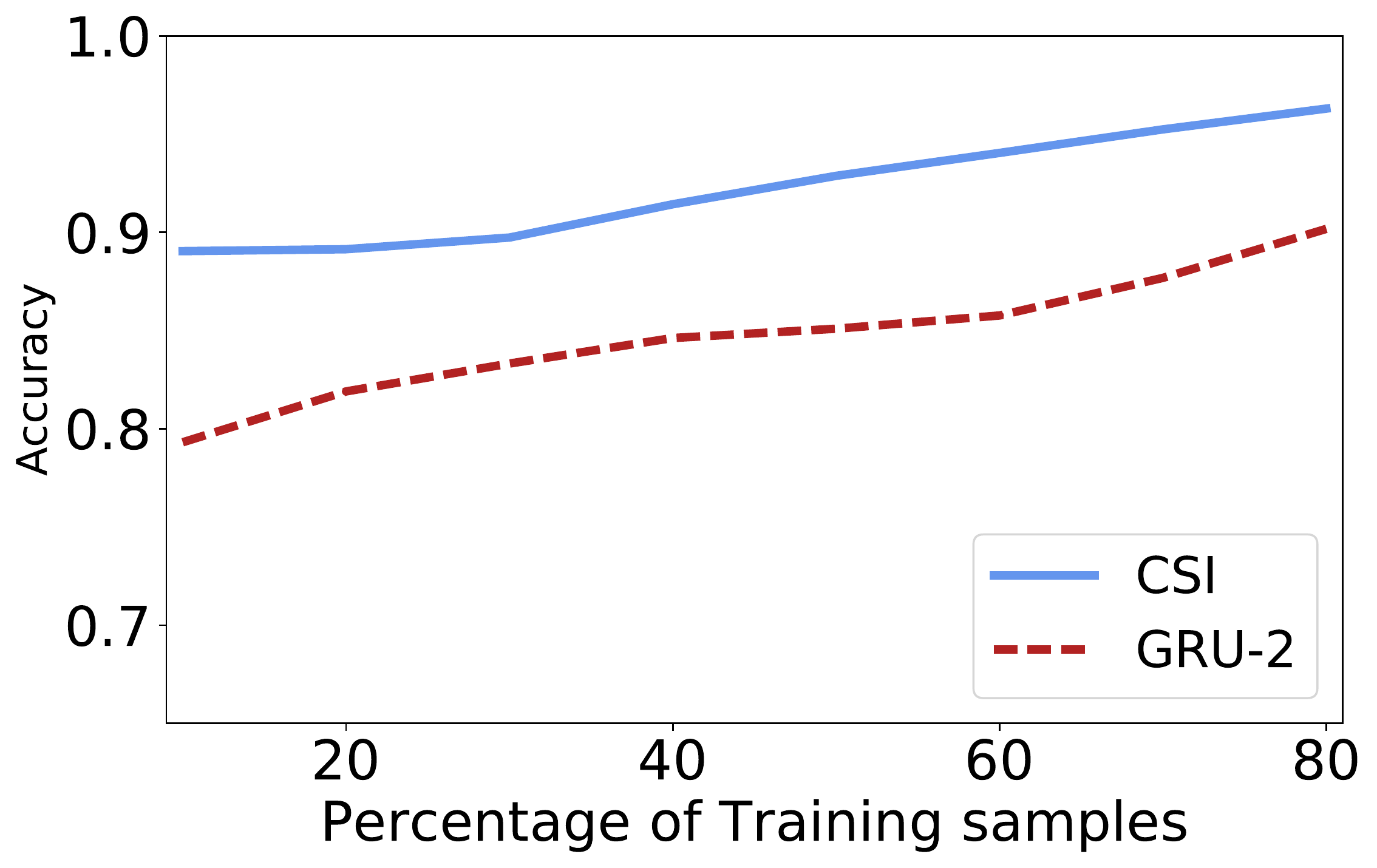}
\caption{{\twitter}}
\end{subfigure}
\hfill
  \begin{subfigure}{0.23\textwidth}
  \centering
    \includegraphics[scale=.19]{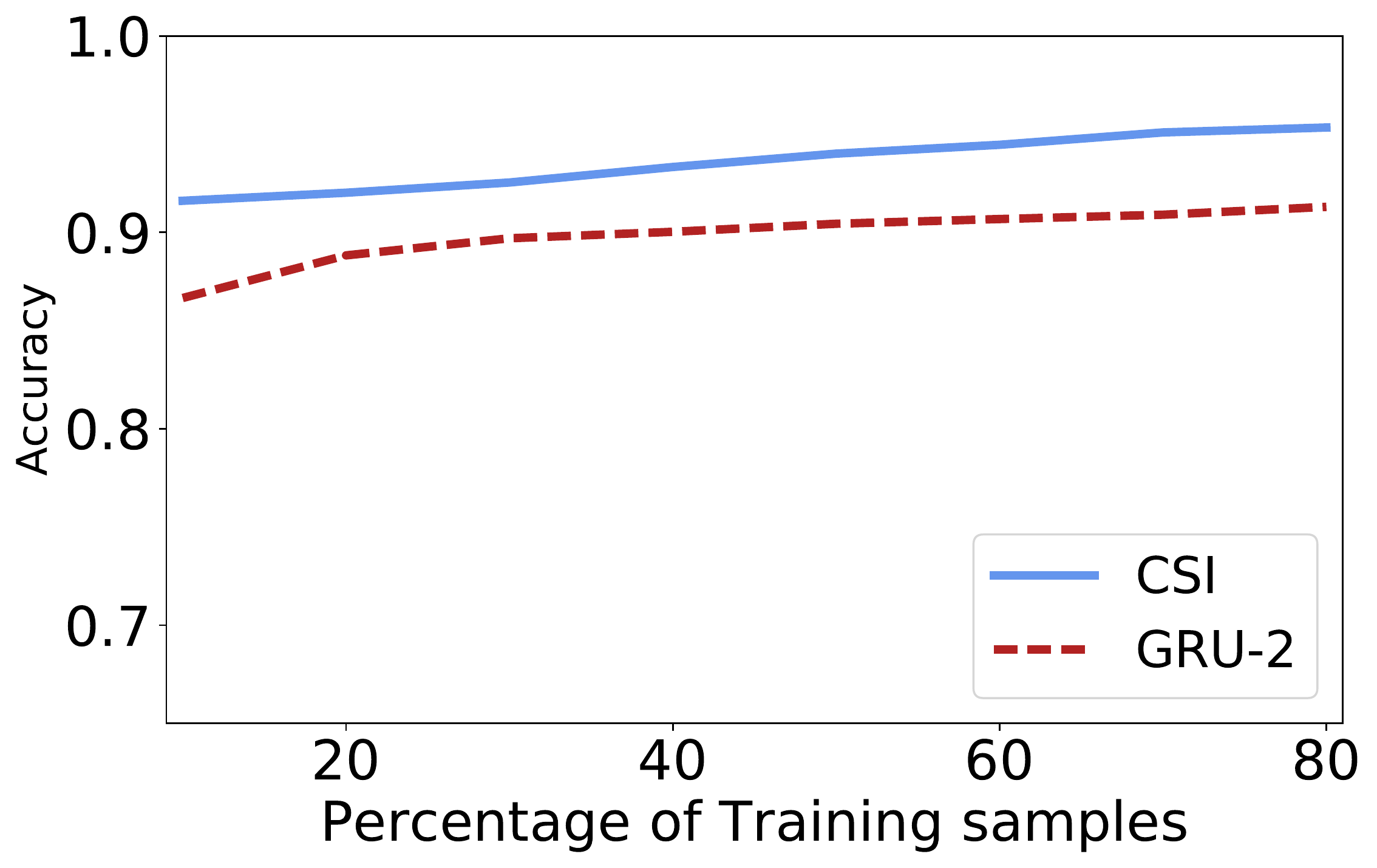}
\caption{{\weibo}}
\end{subfigure}
  \caption{Accuracy vs. the percentage of training samples.\label{fig:less_train}}
  \end{figure}

To study the number of labeled samples {\ourmodel} relies on, we study the accuracy as a function of the training set size. 
Figure~\ref{fig:less_train} shows that even if only 10\% training samples are available, \ourmodel{} can show comparable performance with \textsc{GRU-2}; thus, the {\ourmodel} model is lighter and can be trained more easily with fewer training samples.

\newpage
\subsection{Interpreting user representations}\label{sec:score}
\input{sections/score}

\subsection{Characterizing user behavior}
In this section, we ask whether the users marked as suspicious by {\ourmodel} have any characteristic behavior.
Using the $s_i$ scores of each user we select approximately $25$ users from the most suspicious groups, and the same amount from the least suspicious group.

We consider two properties of user behavior: (1) the \emph{lag} and (2) the \emph{activity}.
To measure lag for each user, we compute the lag in time between time between an article's publication, and when the user first engaged with it.  We then plot the distribution of user lags separated by most and least suspicious, and true and fake news.  Figure~\ref{fig:fa-dist} shows the CDF of the results.
Immediately we see that the most suspicious users in each dataset are some of the first to promote the fake content -- supporting the \emph{source} characteristic.  In contrast, both types of users act similarly on real news.

\begin{figure*}
  \captionsetup[subfigure]{justification=centering}
\begin{subfigure}{0.23\textwidth}
  \centering
    \includegraphics[width=\textwidth]{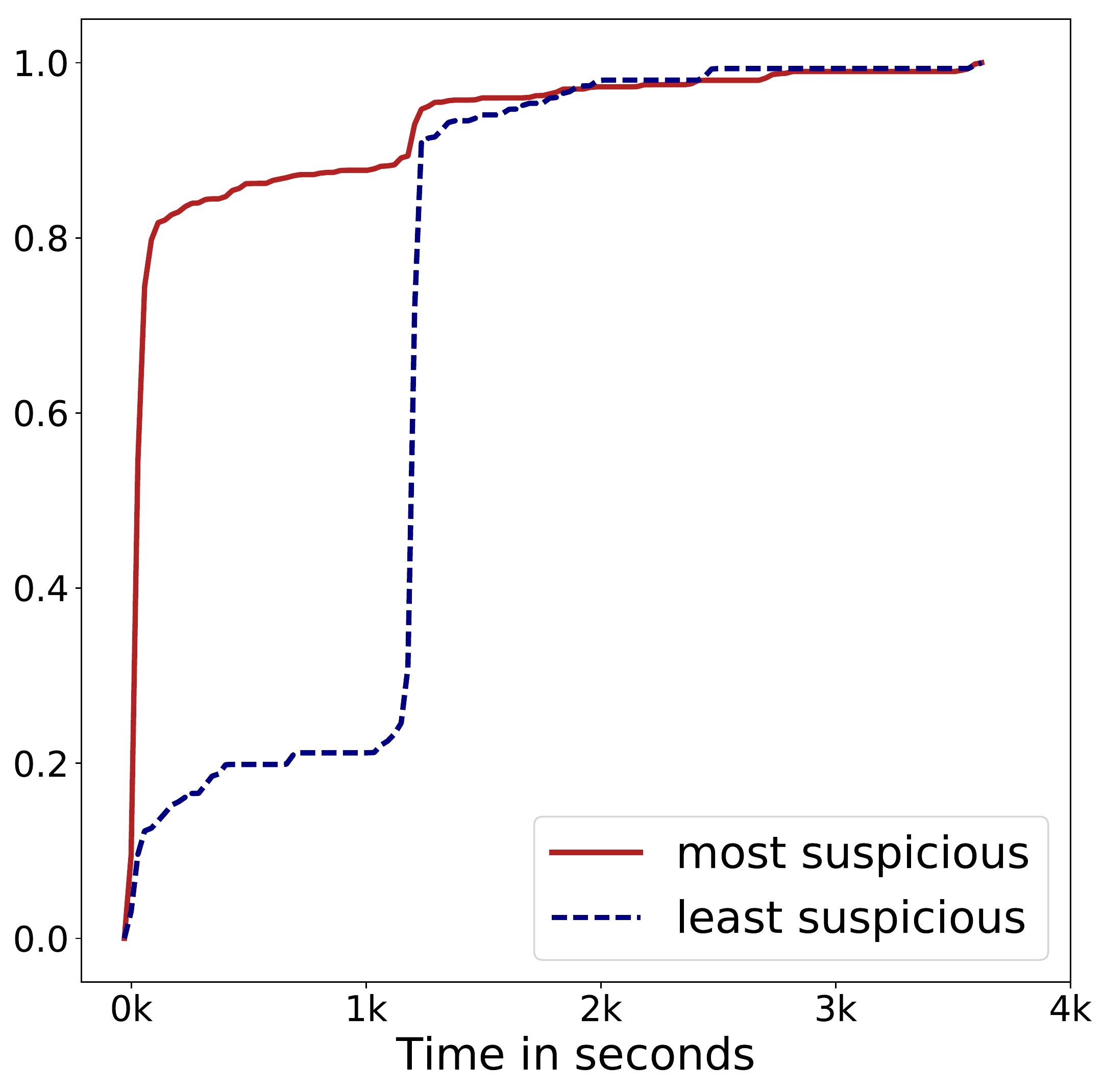}
\caption{Fake news on {\twitter}}
\end{subfigure}
\hspace*{\fill} 
\begin{subfigure}{0.23\textwidth}
  \centering
    \includegraphics[width=\textwidth]{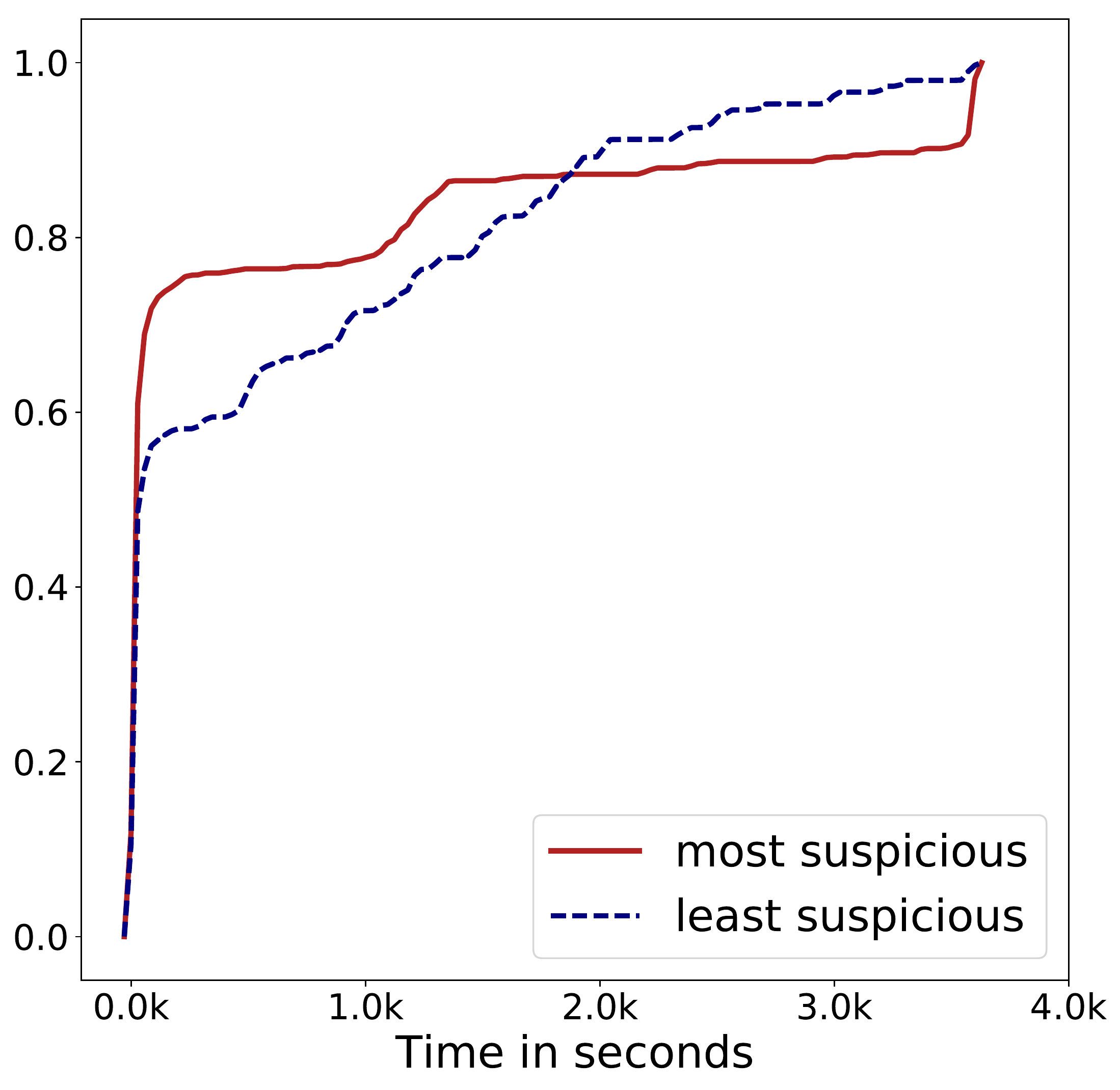}
\caption{True news on {\twitter}}
\end{subfigure}
\hspace{1cm}
\begin{subfigure}{0.23\textwidth}
  \centering
    \includegraphics[width=\textwidth]{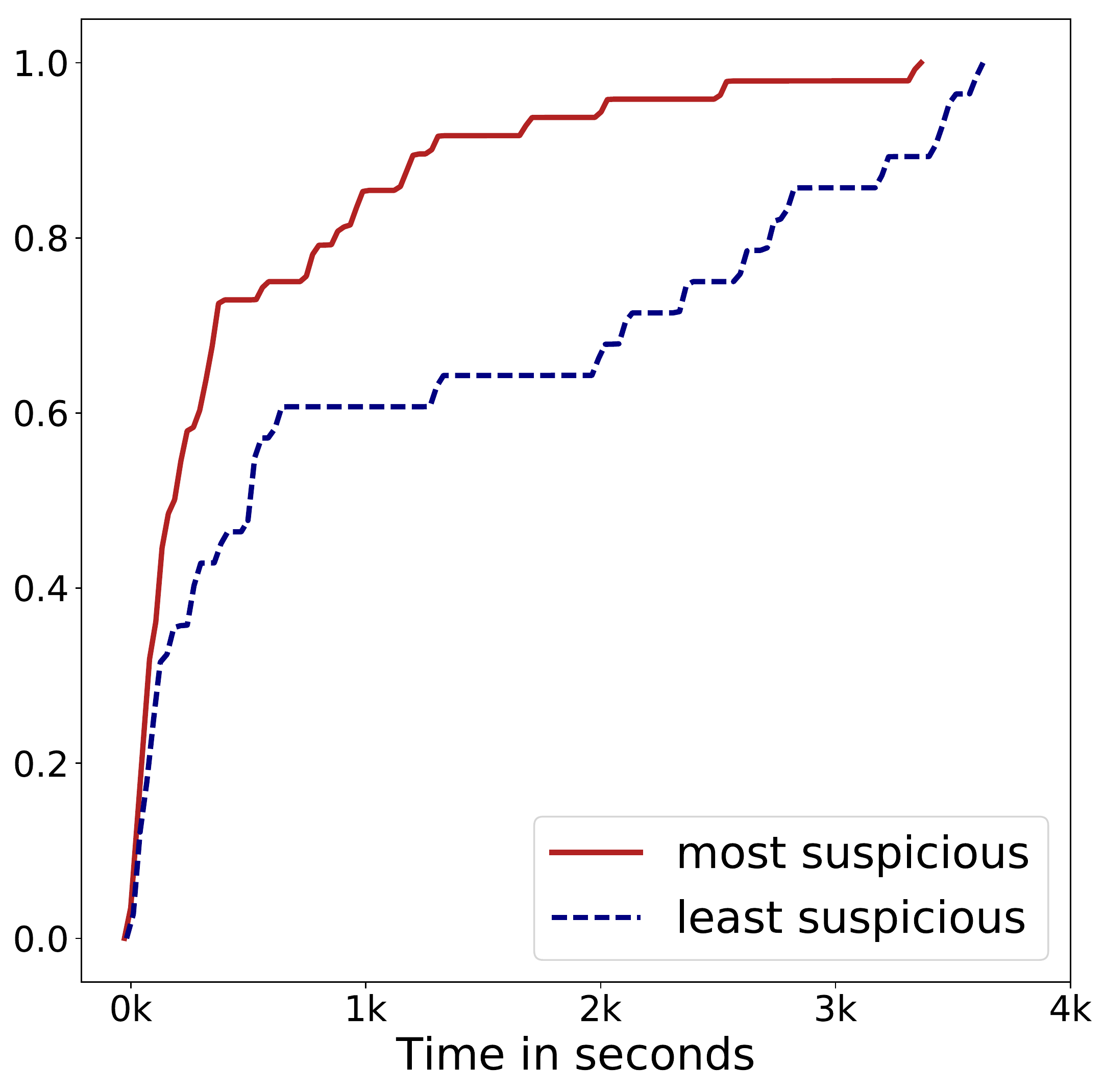}
\caption{Fake news on {\weibo}}
\end{subfigure}
\hspace*{\fill} 
\begin{subfigure}{0.23\textwidth}
  \centering
    \includegraphics[width=\textwidth]{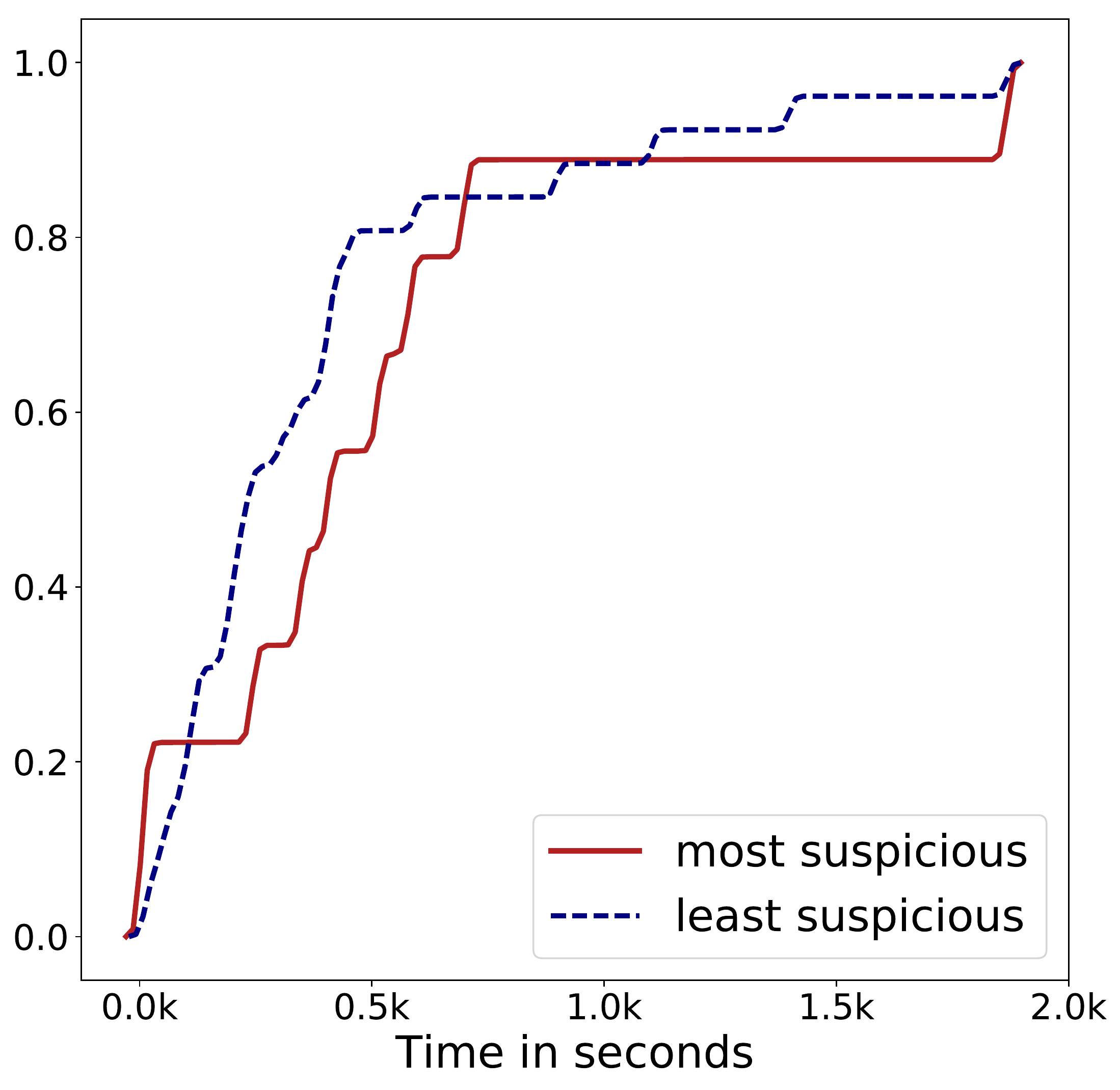}
\caption{True news on {\weibo}}
\end{subfigure}
\caption{Distribution (CDF) of user activity on {\twitter} and {\weibo}.    \label{fig:inter-dist}}
\end{figure*}

Next, we measure the user activity as the time between engagements user $u_i$ had with a particular article $a_j$.  
Figure \ref{fig:inter-dist} 
shows the CDF of user activity.  We see that on both datasets, suspicious users often have bursts of quick engagements with a given article; this behavior differs more significantly from the least suspicious users on fake news than it does on true news.
Interestingly, the behavior of suspicious users on {\twitter} is similar on fake and true news, which may demonstrate a sophistication in fake content promotion techniques.
Overall, these distributions show that the combination of temporal, textual, and user features in $\bb{x}_t$ provides meaningful information to capture the three key characteristics, and for {\ourmodel} to distinguishing suspicious users.

\subsection{Utilizing temporal article representations}
In this section, we investigate the vector $\bb{v}_j$ that is the output of {\modone} for each article $a_j$.  Intuitively, these vectors are a low-dimensional representation of the temporal and textual response an article has received, as well as the types of users the response has come from.  In a general sense, the output of an LSTM has been used for a variety of tasks such as machine translation~\cite{sutskever2014sequence}, question answering~\cite{wang2015long}, and text classification~\cite{lee2016sequential}.  Hence, in the context of this work it is natural to wondering whether these vectors can be used for deeper insight into the space of articles.

\begin{figure}[H]
  \captionsetup[subfigure]{justification=centering}
\begin{subfigure}{0.23\textwidth}
  \centering
        \includegraphics[scale=.25]{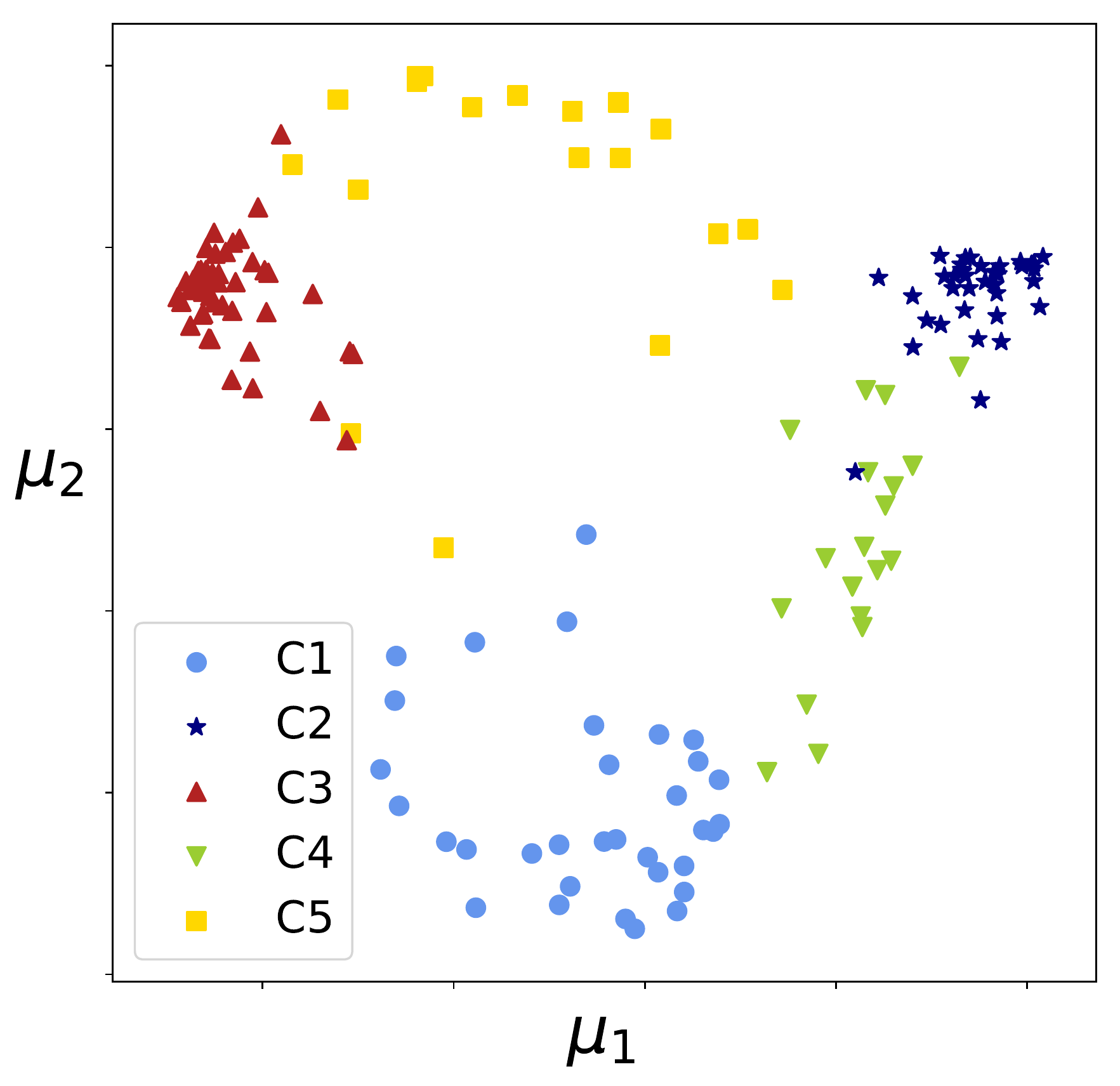}
\caption{{\twitter}}
\end{subfigure}
\hspace*{\fill} 
\begin{subfigure}{0.23\textwidth}
  \centering
        \includegraphics[scale=.25]{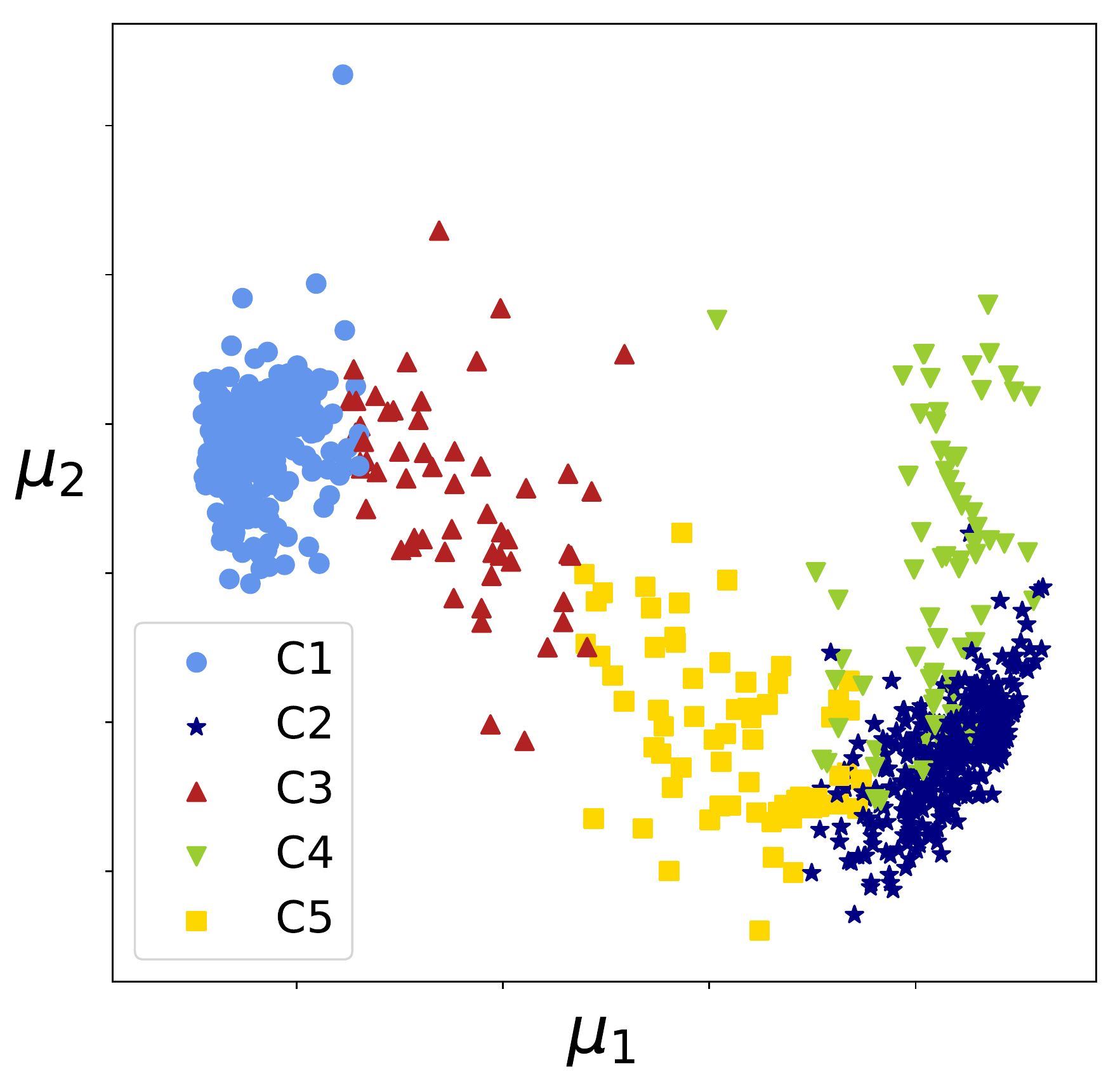}
\caption{{\weibo}}
\end{subfigure}
\caption{Article clustering with {\articlevec} on {\twitter} and {\weibo}.    \label{fig:clustboth}}
\end{figure}

As an example, we consider applying Spectral Clustering for a more fine-grained partition than two classes.  We consider the set of $\bb{v}_j$ associated with the test set of {\twitter} and {\weibo} articles,  and set $k=5$ clusters according to the elbow curve.  Figure~\ref{fig:clustboth} shows the results in the space of the first two singular vectors ($\mu_1$ and $\mu_2$) of the matrix formed by the vectors $\bb{v}_j$ for each respective dataset, with one color for each cluster.

Table~\ref{tab:cluststats} shows the breakdown of true and false articles in each cluster.  We can see that the results gives a natural division both among true and fake articles.  For example, on the {\twitter} datasets, while both C2 and C4 are composed of mostly fake news, we can see that the projections of their temporal representation are quite separated.  This separation suggests that there may be different types of fake news which exhibit slightly different signals in the \emph{text}, \emph{response}, and \emph{source} characteristics, for example, satire and spam.
The {\weibo} data shows two poles: C1 in the top left corresponds largely to true news, while C2 and C4 captures different types of fake news.  Meanwhile, C3 and C5 which are spread across the middle, have more mixed membership.

In the context of  the general framework described in Section~\ref{sec:model}, the results show that the $\bb{v}_j$ vectors produced by the {\modone} module offer insight into the population of users with respect to their behavior towards fake news.  Aside from the classification output of the model, the representations can be used stand-alone for gaining insight about targets (articles) in the data.

\begin{table}
\begin{tabular}{ c  c  c  c  c  c  c }
\toprule
         \multicolumn{3}{c}{{\twitter}}      && \multicolumn{3}{c}{{\weibo}}      \\ 
        \cmidrule(lr){1-3}   \cmidrule(lr){5-7}
Cluster & True & False && Cluster & True & False \\ \midrule
1 &16 &17 & & 1 &362& 5\\
2 & 5&33 &&   2 & 16& 326\\
3 & 46&2 &  & 3 & 45& 10\\
4 & 3&16 &&   4 & 0& 72\\
5 & 11&8 &  & 5 & 28& 37\\
\bottomrule
\end{tabular}
\caption{Cluster statistics for {\twitter} and {\weibo} for Figure~\ref{fig:clustboth}.\label{tab:cluststats}}
\end{table}

%% file: sections/score.tex

\nredit{
In this section, we analyze the output of {\modtwo} which is a score $\userscore$ and a representation $\uservec$ for every user.  Since the available data does not have ground-truth labels on users, we perform a qualitative evaluation of the information contained in $(\userscore,\uservec)$ with respect to the \emph{source} characteristic of fake news.

Although we lack user-labels, the dataset still contains information that can be used as a proxy.  In particular, we want to evaluate whether $(\userscore,\uservec)$ captures the suspicious behavior of users in terms promotion of fake news and group behavior.  For the former, a reasonable proxy is the fraction of fake news a user engages with, denoted $\ell_i\in [0,1]$ with $0.0$ meaning the user has never reacted to fake news, and $1.0$ meaning the engagements are exclusively with fake news.  } 
 In addition, we consider the corresponding scores for articles as the average over users, namely $\scoreart$ is the average of $\userscore$  and $\lambda_j$ is the average of $\ell_i$ over $u_i$ that engaged with $a_j$.

\begin{figure}[H]
  \captionsetup[subfigure]{justification=centering}
\begin{subfigure}{0.23\textwidth}
  \centering
    \includegraphics[scale=.23]{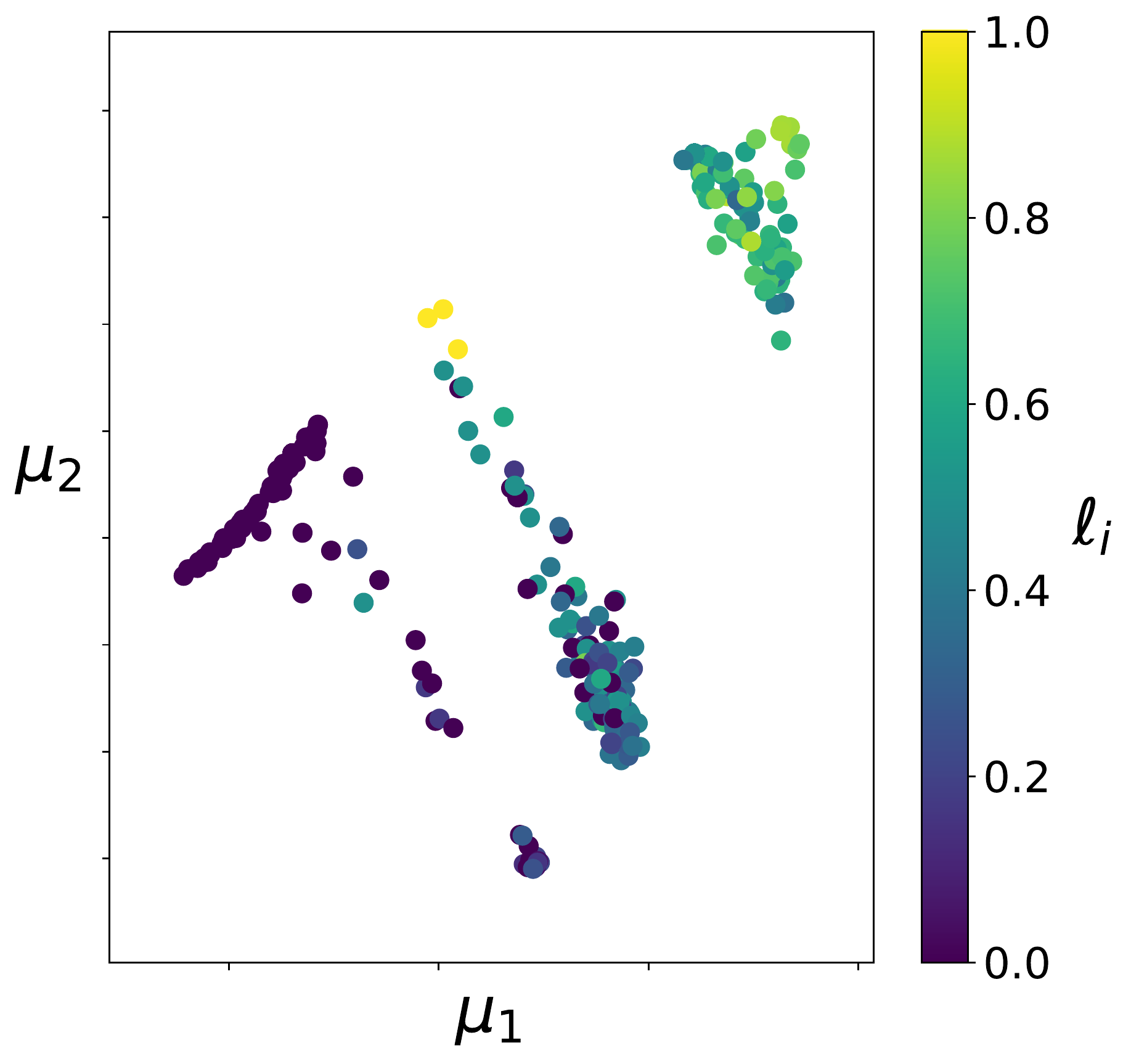}
\caption{{\twitter} users}
\end{subfigure}
\hfill
\begin{subfigure}{0.23\textwidth}
  \centering
    \includegraphics[scale=.23]{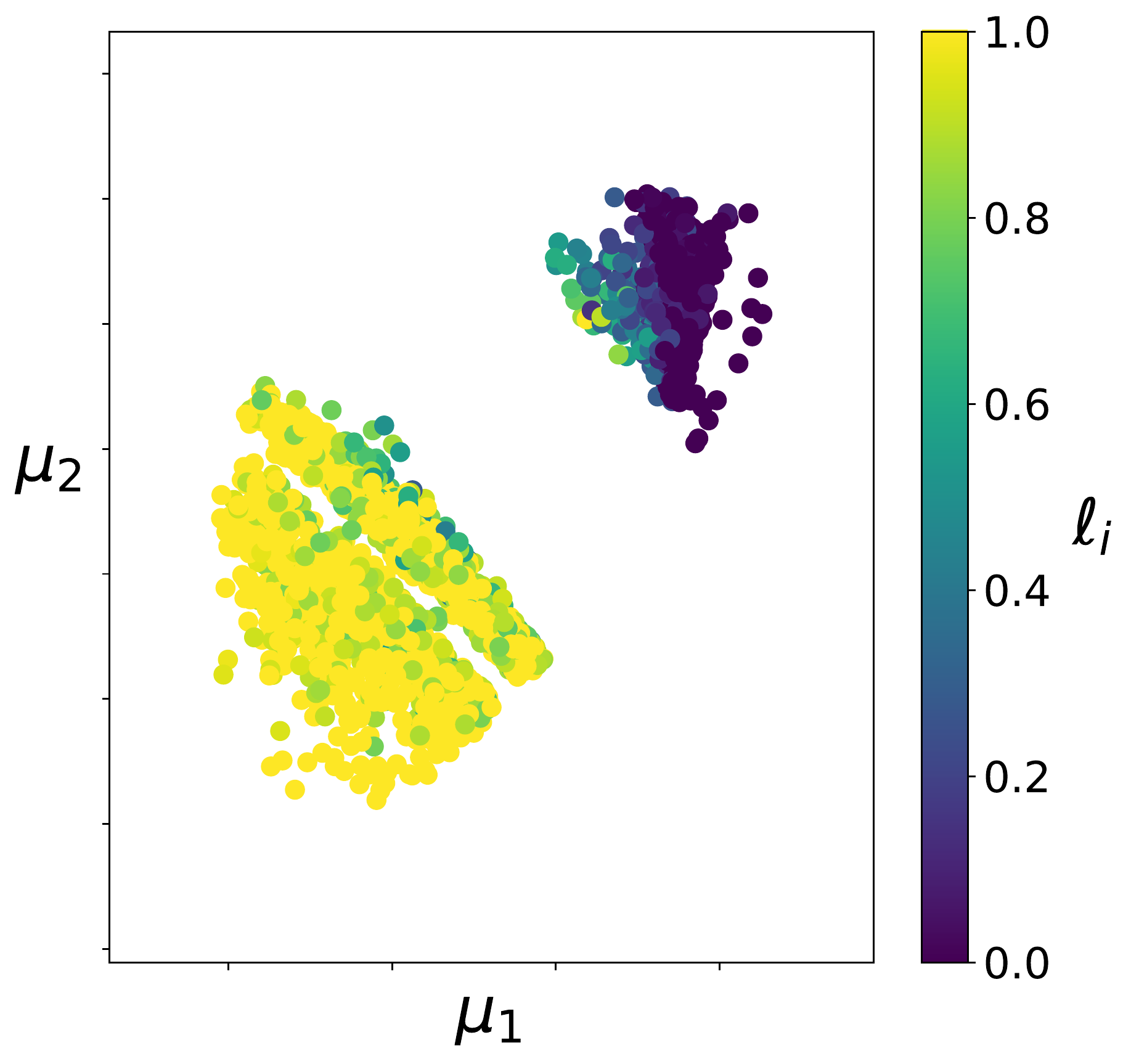}
\caption{{\weibo} users}
\end{subfigure}
\caption{Projection of user vectors $\bb{z}_j$.\label{fig:user_proj}}
\end{figure}

\nredit{
To test the extent to which $(\userscore,\uservec)$ capture $\ell_i$, we compute the correlation between the two measures across users; Table~\ref{tab:scoresreg} shows the Pearson correlation coefficient and significance.  For both datasets and on both sides of the user-article engagement, we find a statistically significant positive relationship between the two scores.  Results are consistent for the Spearman coefficient and for ordinary least squares regression(OLS).  In addition, Figures~\ref{fig:zscore-twitter} and \ref{fig:zscore-weibo} show the distribution of $\ell_i$ among a subset of users with highest and lowest $\userscore$.  
Most of the users who were assigned a high $\userscore$ by {\ourmodel} (marked as most suspicious) have $\ell_i$ close to $1$, while those with low $\userscore$ have low $\ell_i$.  Altogether, the results demonstrate that $\userscore$ and $\scoreart$ hold meaningful information with respect to user levels of engagement with fake news.

\begin{table}[H]
\centering
\begin{tabular}{l c c }
\toprule
      \multicolumn{1}{c}{}     & \multicolumn{1}{c}{User}      & \multicolumn{1}{c}{Article}                   \\ 
      \midrule
{\twitter} & $0.525$***          &      $0.671$***       \\ 
{\weibo}    & $0.485$***            & $0.646$***                    \\ \bottomrule
\end{tabular}
\caption{Correlation between $\ell_i$ and $\uservec$ with statistical significance as *$<0.1$, **$<0.05$, and ***$ < 0.01$.\label{tab:scoresreg} }
\end{table}

\begin{figure}[t]
  \captionsetup[subfigure]{justification=centering}
\begin{subfigure}{0.23\textwidth}
  \centering
\includegraphics[height=4.2cm, width=4.2cm]{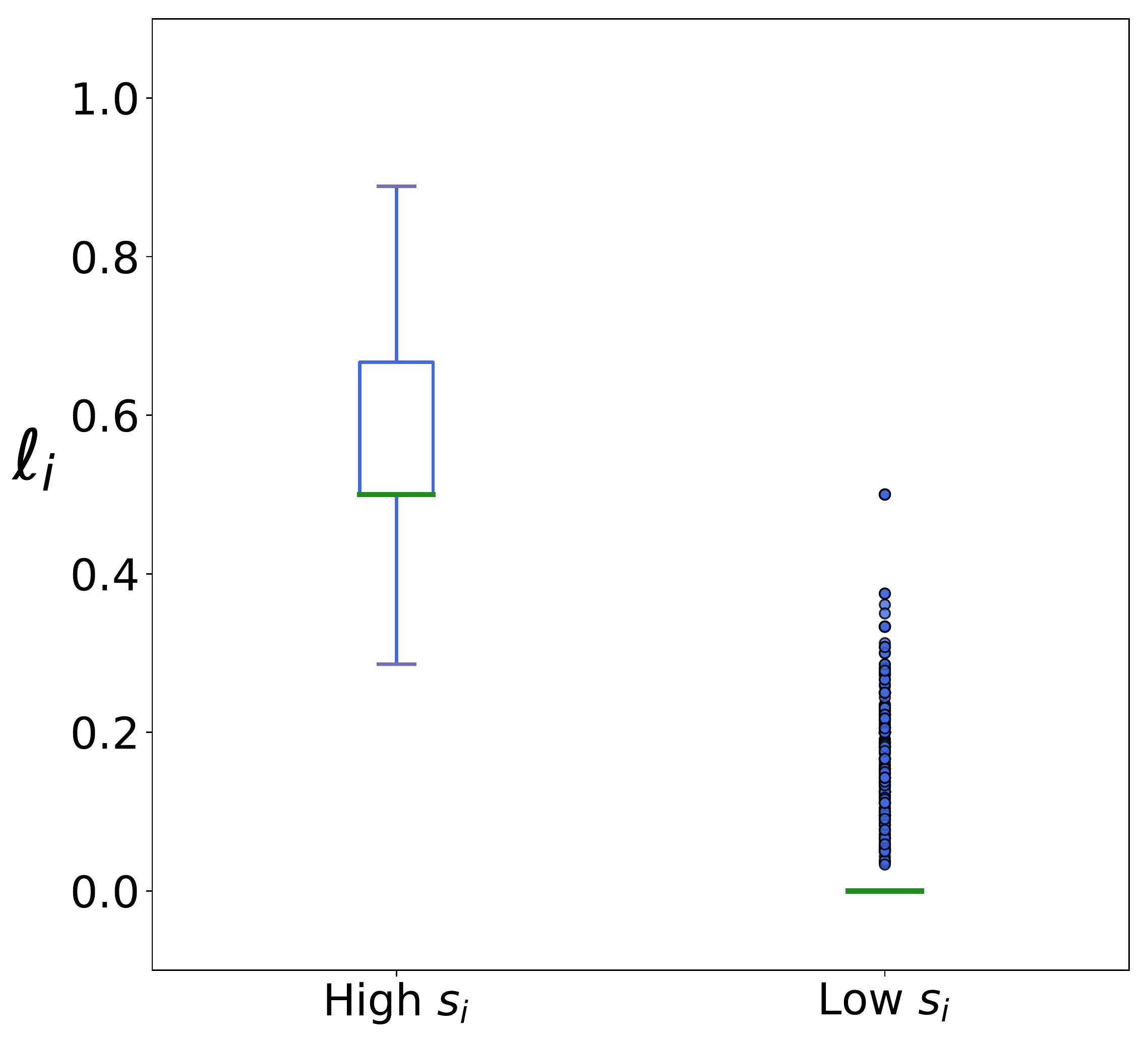}
\caption{{\twitter}\label{fig:zscore-twitter}}
\end{subfigure}
\hspace*{\fill} 
\begin{subfigure}{0.23\textwidth}
  \centering
  \includegraphics[height=4.2cm, width=4.2cm]{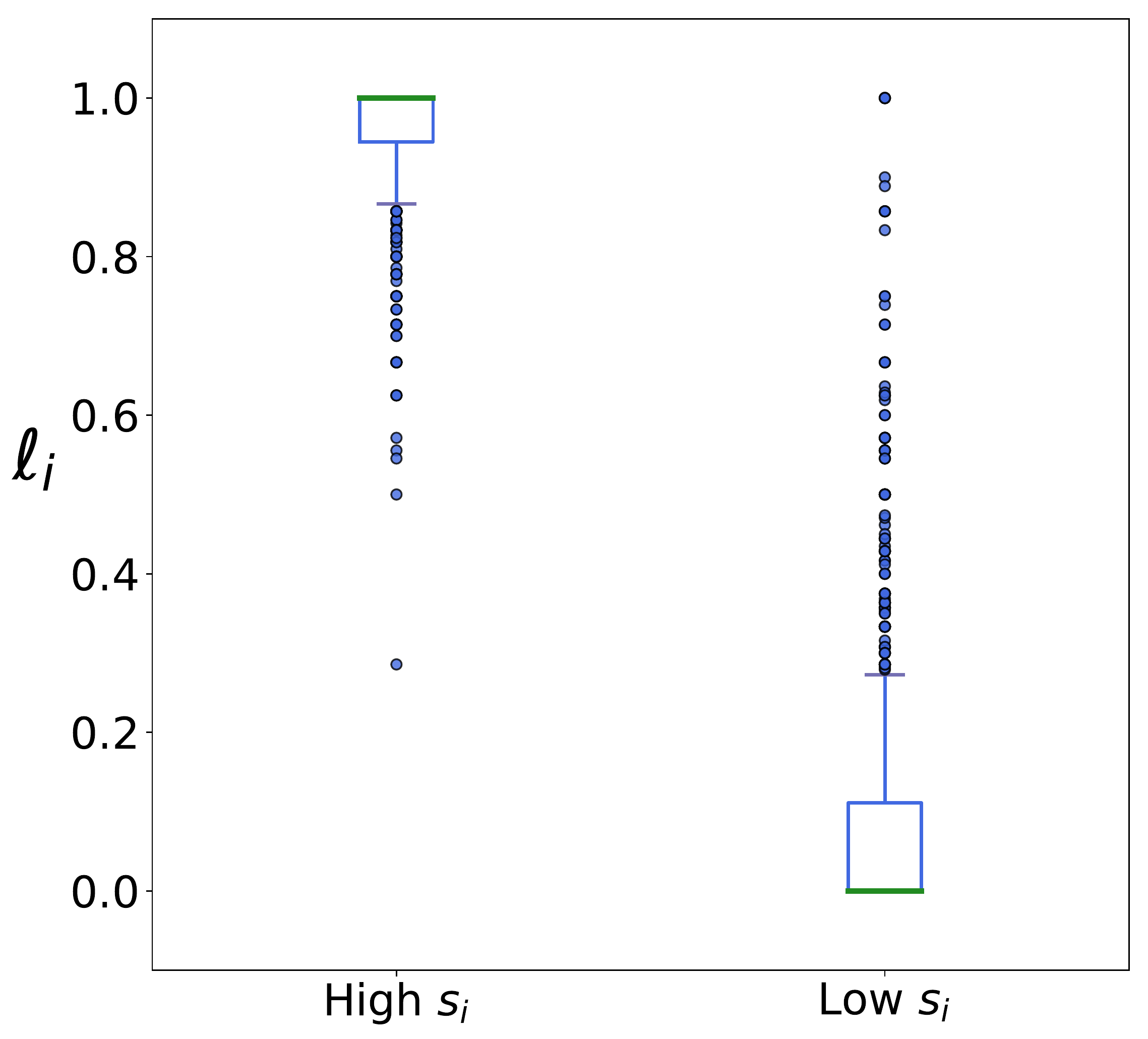}
\caption{{\weibo}\label{fig:zscore-weibo}}
\end{subfigure}
\caption{ Distribution of $\ell_i$ over users marked as high and low suspicion according to the $\bb{s}$ vector produce by {\ourmodel}.}
\end{figure}

\newpage
To investigate the relation of $\uservec$ to $\ell_i$, we regress the cosine distance between $\uservec$ and $\userveci{i'}$ against the difference between $\ell_i$ and $\ell_{i'}$ for each pair of users $(i,i')$.  Consistent with results for $\userscore$, we find a positive correlation of $0.631$ for {\twitter} and $0.867$ for {\weibo}, both of which are statistically significant  at the $1\%$ level.  Further, we visualize the space of user representations by projecting a sample of the vectors $\uservec$ onto the first and second singular vectors $\mu_1$ and $\mu_2$ of the matrix of $\tilde{\bb{y}_i}$'s.  Figure~\ref{fig:user_proj} shows the projection for both datasets, where each point corresponds to a user $u_i$ and is colored according to $\ell_i$.  We see that the space exhibits a strong separation between users with extreme $\ell_i$, suggesting that the vectors $\uservec$ offer a good latent representation of user behavior with respect to fake news and can be used for deeper user analysis.}

\begin{figure*}[t]
  \captionsetup[subfigure]{justification=centering}
\begin{subfigure}{0.23\textwidth}
  \centering
    \includegraphics[width=\textwidth]{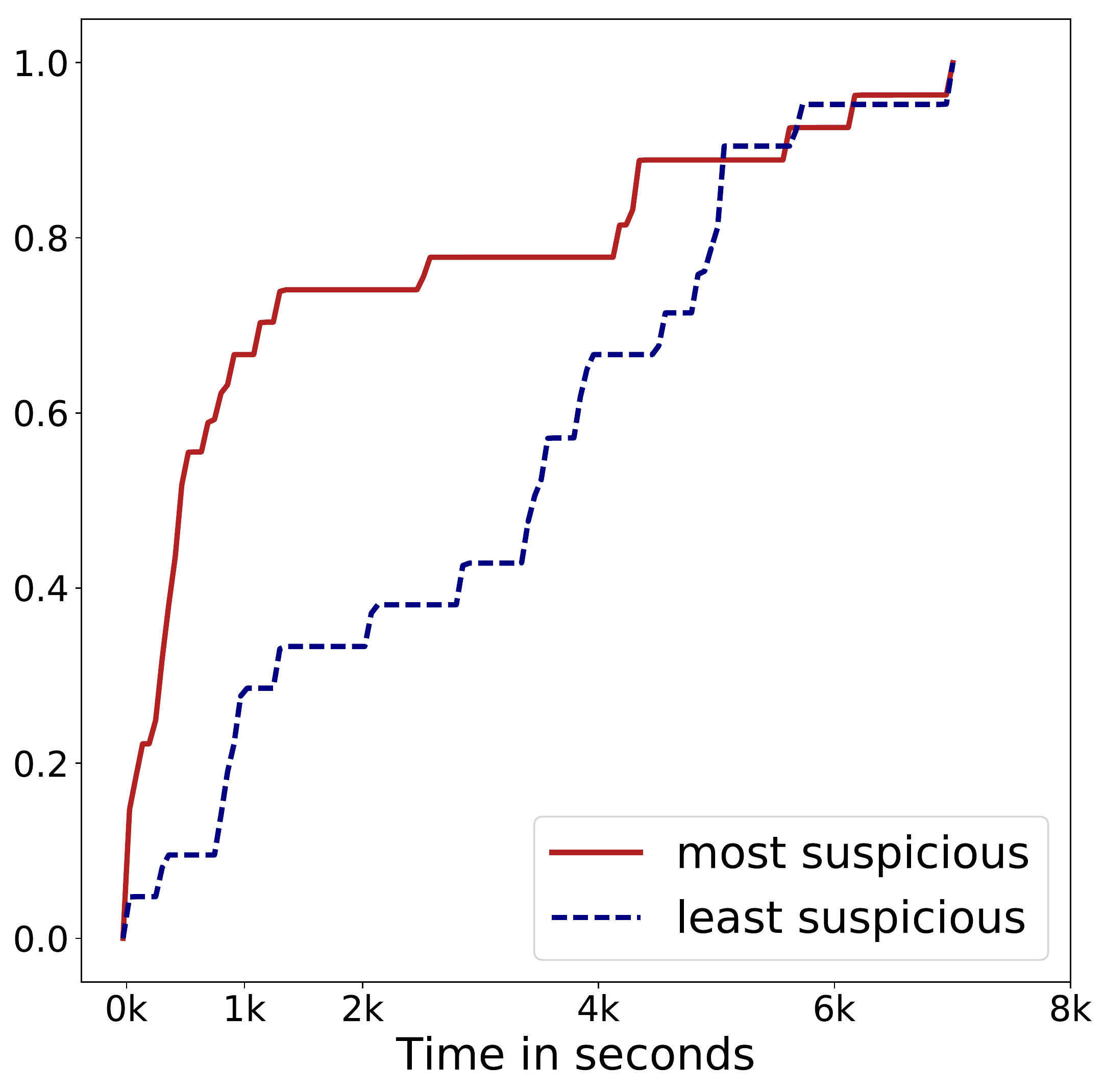}
\caption{Fake news on {\twitter}}
\end{subfigure}
\hspace*{\fill} 
\begin{subfigure}{0.23\textwidth}
  \centering
    \includegraphics[width=\textwidth]{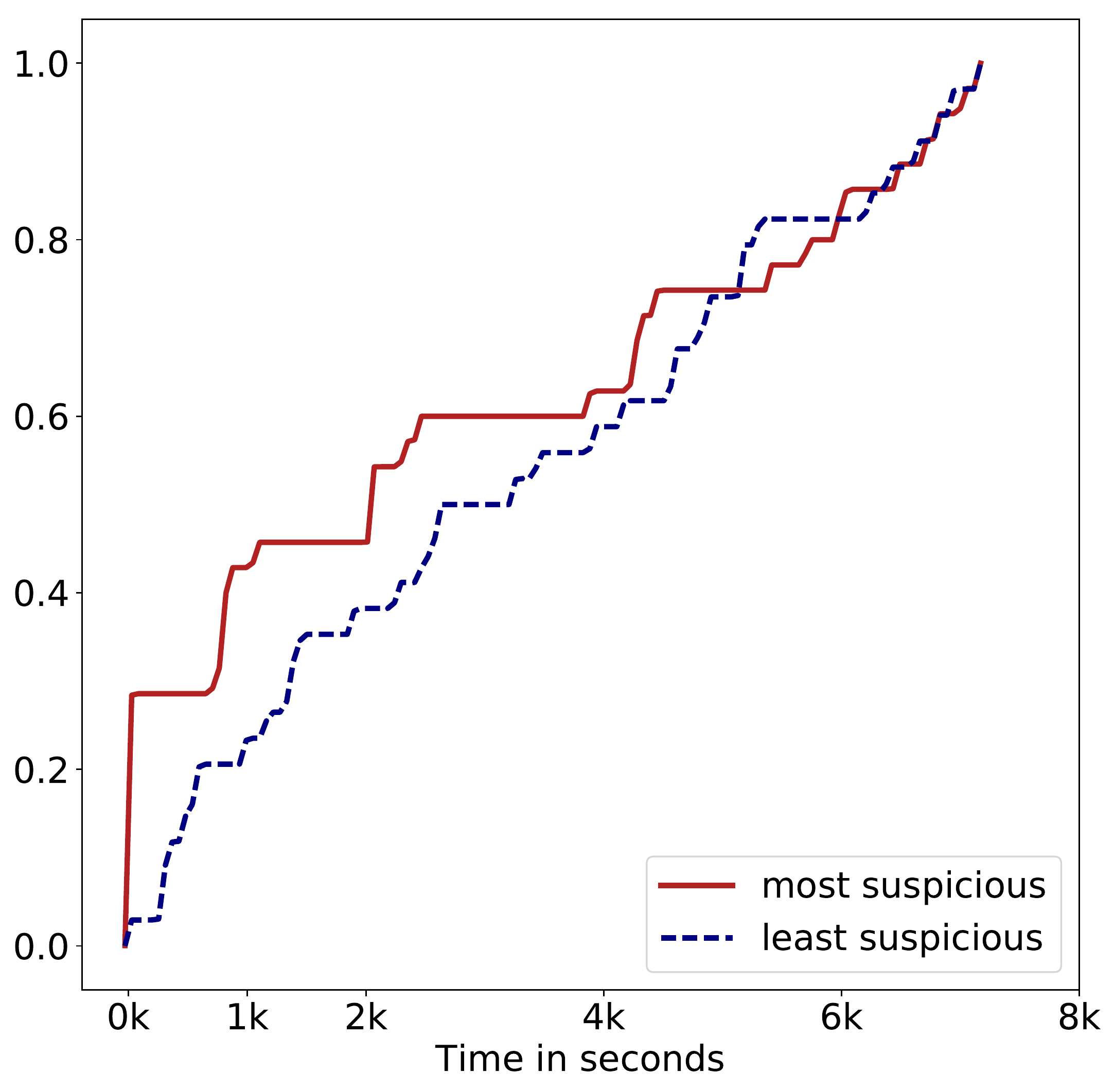}
\caption{True news on {\twitter}}
\end{subfigure}
\hspace*{1cm} 
\begin{subfigure}{0.23\textwidth}
  \centering
    \includegraphics[width=\textwidth]{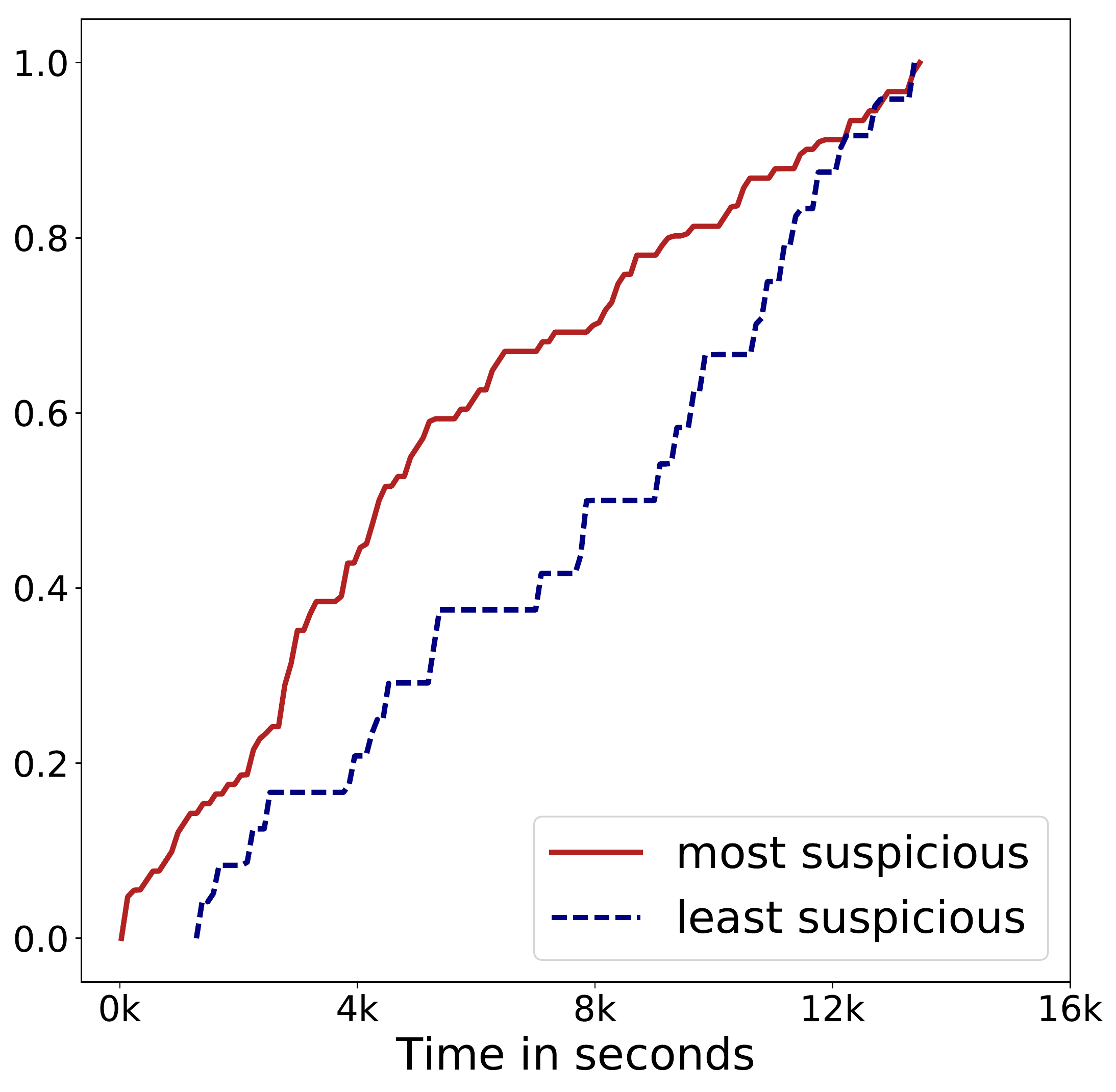}
\caption{Fake news on {\weibo}}
\end{subfigure}
\hspace*{\fill} 
\begin{subfigure}{0.23\textwidth}
  \centering
    \includegraphics[width=\textwidth]{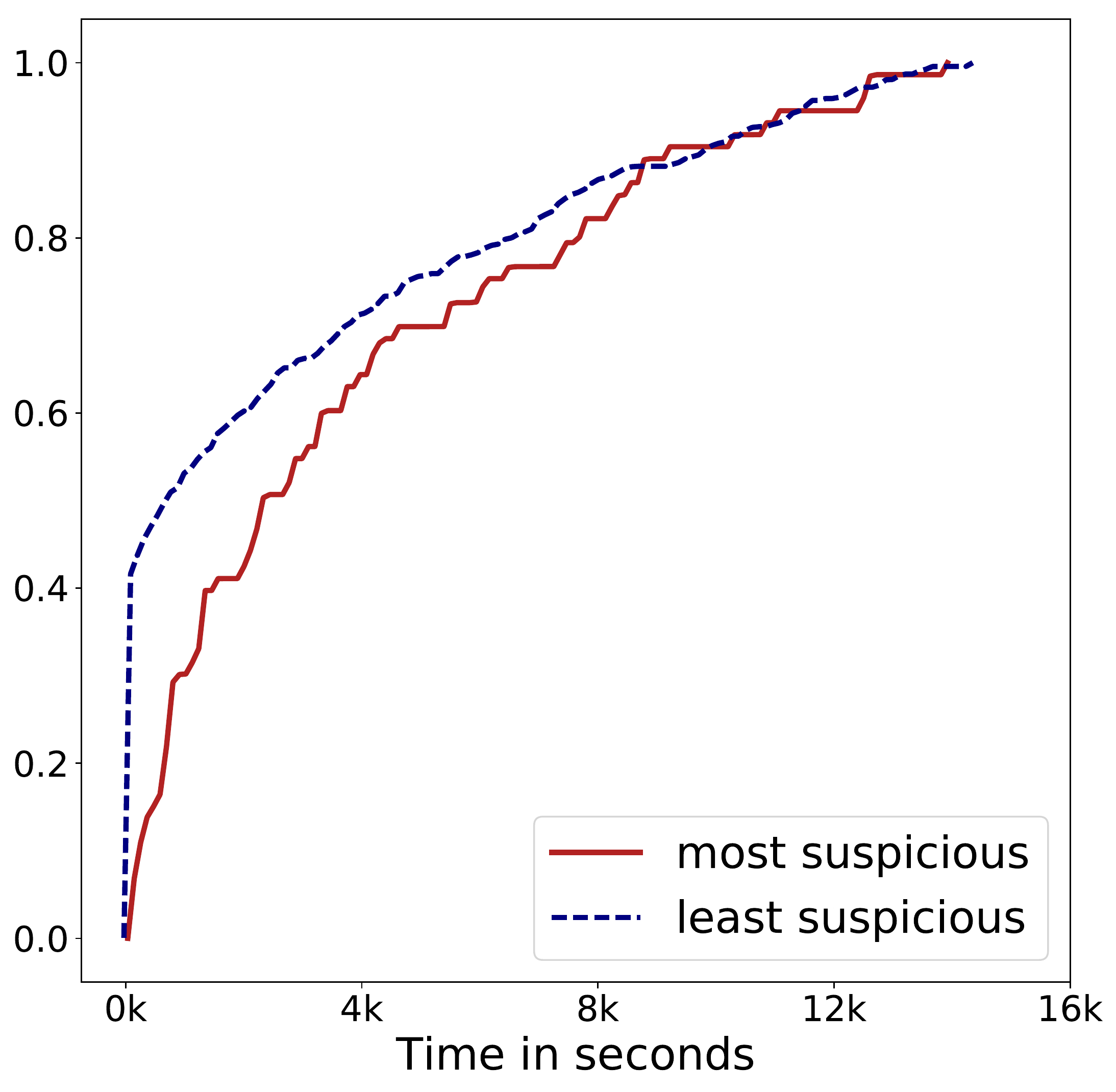}
\caption{True news on {\weibo}}
\end{subfigure}
\caption{Distribution (CDF) of user lags on {\twitter} and {\weibo}.  \label{fig:fa-dist}}
\end{figure*}

Next, we analyze the propensity of $(\userscore,\uservec)$ to capture group behavior.  We construct an implicit user graph by adding an edge between users who have engaged with the same article, and by analyze the clustering of users in the graph. 
We apply the BiMax algorithm proposed by Preli{\'c} \etal{}~\cite{prelic2006systematic} to search for biclusters in the adjacency matrix.\footnote{BiMax available here http://www.kemaleren.com/the-bimax-algorithm.html}
We find that for both datasets, users with large $\ell_i$  participate in more and larger biclusters than those with low $\ell_i$.  Further, biclusters for users with large $\ell_i$ are formed largely with fake news articles, while those for low $\ell_i$ are largely with true news. 

This suggests that suspicious users exhibit the \emph{source} characteristic with respect to fake news.   In addition, for each pair of users $(u_i,u_{i'})$ we compute the Jaccard distance between the set of articles they interacted with.  We compute the correlation between this quantity and $|\userscore-s_{i'}|$ as well as the cosine distance between $\uservec$ and $\userveci{i'}$.  For the former we find a correlation of $0.36$ for {\twitter} and $0.21$ for {\weibo}, and for the latter we find $0.30$ for {\twitter} and $0.16$ for {\weibo}.  All results are significant at the $1\%$ level, with Spearman correlation and OLS giving consistent results.

Overall, despite lack of ground-truth labels on users, our analysis demonstrates that the {\modtwo} module captures meaningful information with respect to the the \emph{source} characteristic.  The user score $\userscore$ provides the model with an indication of the suspiciousness of user $u_i$ with respect to group behavior and fake news engagement.  Further, the $\uservec$ vector provides a representation of each user that can be used for deeper analysis of user behavior in the data.


%% file: sections/conclusion.tex

In this work, we study the timely problem of fake news detection.  While existing work has typically addressed the problem by focusing on either the \emph{text}, the \emph{response} an article receives, or the users who \emph{source} it, we argue that it is important to incorporate all three.
 We propose the {\ourmodel} model which is composed of three modules.  The first module, {\modone}, captures the abstract temporal behavior of user encounters with articles, as well as temporal textual and user features, to measure \emph{response} as well as the \emph{text}.  The second component, {\modtwo}, estimates a \emph{source} suspiciousness score for every user, which is then combined with the first module by {\modthree} to produce a predicted label for each article.  

 The separation into modules allows {\ourmodel} to output a prediction separately on users and articles, incorporating each of the three characteristics,  meanwhile combining the information for classification.  Experiments on two real-world datasets demonstrate the accuracy of {\ourmodel} in classifying fake news articles.  Aside from accurate prediction, the {\ourmodel} model also produces latent representations of both users and articles that can be used for separate analysis; we demonstrate the utility of both the extracted representations and the computed user scores.  

The {\ourmodel} model is general in that it does not make assumptions on the distribution of user behavior, on the particular textual context of the data, nor on the underlying structure of the data.  Further, by utilizing the power of neural networks, we incorporate different sources of information, and capture the temporal evolution of engagements from both parties, users and articles.  At the same time, the model allows for easy incorporation of richer data, such as user profile information, or advanced text libraries.
Overall our work demonstrates the value in modeling the three intuitive and powerful characteristics of fake news.

Despite encouraging results, fake news detection remains a challenging problem with many open questions.
One particularly interesting direction would be to build models that incorporate concepts from reinforcement learning and crowd sourcing.  Including humans in the learning process could lead to more accurate and, in particular, more timely predictions.

%% file: main.bbl

\begin{thebibliography}{00}


\ifx \showCODEN    \undefined \def \showCODEN     #1{\unskip}     \fi
\ifx \showDOI      \undefined \def \showDOI       #1{{\tt DOI:}\penalty0{#1}\ }
  \fi
\ifx \showISBNx    \undefined \def \showISBNx     #1{\unskip}     \fi
\ifx \showISBNxiii \undefined \def \showISBNxiii  #1{\unskip}     \fi
\ifx \showISSN     \undefined \def \showISSN      #1{\unskip}     \fi
\ifx \showLCCN     \undefined \def \showLCCN      #1{\unskip}     \fi
\ifx \shownote     \undefined \def \shownote      #1{#1}          \fi
\ifx \showarticletitle \undefined \def \showarticletitle #1{#1}   \fi
\ifx \showURL      \undefined \def \showURL       #1{#1}          \fi
\providecommand\bibfield[2]{#2}
\providecommand\bibinfo[2]{#2}
\providecommand\natexlab[1]{#1}
\providecommand\showeprint[2][]{arXiv:#2}

\bibitem[\protect\citeauthoryear{globalvoices.org}{kre}{2015}]%
        {kremlin}
 \bibinfo{year}{2015}\natexlab{}.
\newblock \bibinfo{title}{Social Network Analysis Reveals Full Scale of
  Kremlin's Twitter Bot Campaign.}
\newblock   (\bibinfo{date}{April} \bibinfo{year}{2015}).
\newblock
\showURL{%
\url{globalvoices.org/2015/04/02/analyzing-kremlin-twitter-bots/}}


\bibitem[\protect\citeauthoryear{npr.org}{npr}{2016}]%
        {npr}
 \bibinfo{year}{2016}\natexlab{}.
\newblock \bibinfo{title}{Students Have 'Dismaying' Inability To Tell Fake News
  From Real, Study Finds.}
\newblock   (\bibinfo{date}{November} \bibinfo{year}{2016}).
\newblock
\showURL{%
\url{www.npr.org/sections/thetwo-way/2016/11/23/503129818/study-finds-students-have-dismaying-inability-to-tell-fake-news-from-real}}


\bibitem[\protect\citeauthoryear{theguardian.com}{ger}{2017}]%
        {german}
 \bibinfo{year}{2017}\natexlab{}.
\newblock \bibinfo{title}{Germany investigating unprecedented spread of fake
  news online.}
\newblock   (\bibinfo{date}{January} \bibinfo{year}{2017}).
\newblock
\showURL{%
\url{www.theguardian.com/world/2017/jan/09/germany-investigating-spread-fake-news-online-russia-election}}


\bibitem[\protect\citeauthoryear{Beutel, Xu, Guruswami, Palow, and
  Faloutsos}{Beutel et~al\mbox{.}}{2013}]%
        {beutel2013copycatch}
\bibfield{author}{\bibinfo{person}{Alex Beutel}, \bibinfo{person}{Wanhong Xu},
  \bibinfo{person}{Venkatesan Guruswami}, \bibinfo{person}{Christopher Palow},
  {and} \bibinfo{person}{Christos Faloutsos}.} \bibinfo{year}{2013}\natexlab{}.
\newblock \showarticletitle{Copycatch: stopping group attacks by spotting
  lockstep behavior in social networks}. In \bibinfo{booktitle}{{\em
  Proceedings of the 22nd international conference on World Wide Web}}. ACM,
  \bibinfo{pages}{119--130}.
\newblock


\bibitem[\protect\citeauthoryear{Castillo, El-Haddad, Pfeffer, and
  Stempeck}{Castillo et~al\mbox{.}}{2014}]%
        {castillo2014characterizing}
\bibfield{author}{\bibinfo{person}{Carlos Castillo}, \bibinfo{person}{Mohammed
  El-Haddad}, \bibinfo{person}{J{\"u}rgen Pfeffer}, {and} \bibinfo{person}{Matt
  Stempeck}.} \bibinfo{year}{2014}\natexlab{}.
\newblock \showarticletitle{Characterizing the life cycle of online news
  stories using social media reactions}. In \bibinfo{booktitle}{{\em
  Proceedings of the 17th ACM conference on Computer supported cooperative work
  \& social computing}}. ACM, \bibinfo{pages}{211--223}.
\newblock


\bibitem[\protect\citeauthoryear{Castillo, Mendoza, and Poblete}{Castillo
  et~al\mbox{.}}{2011}]%
        {castillo2011information}
\bibfield{author}{\bibinfo{person}{Carlos Castillo}, \bibinfo{person}{Marcelo
  Mendoza}, {and} \bibinfo{person}{Barbara Poblete}.}
  \bibinfo{year}{2011}\natexlab{}.
\newblock \showarticletitle{Information credibility on twitter}. In
  \bibinfo{booktitle}{{\em Proceedings of the 20th international conference on
  World wide web}}. ACM, \bibinfo{pages}{675--684}.
\newblock


\bibitem[\protect\citeauthoryear{Chavoshi, Hamooni, and Mueen}{Chavoshi
  et~al\mbox{.}}{2016}]%
        {chavoshidebot}
\bibfield{author}{\bibinfo{person}{Nikan Chavoshi}, \bibinfo{person}{Hossein
  Hamooni}, {and} \bibinfo{person}{Abdullah Mueen}.}
  \bibinfo{year}{2016}\natexlab{}.
\newblock \showarticletitle{DeBot: Twitter Bot Detection via Warped
  Correlation}.
\newblock \bibinfo{journal}{{\em 2016 IEEE 16th International Conference on
  Data Mining (ICDM)\/}} (\bibinfo{year}{2016}), \bibinfo{pages}{817--822}.
\newblock


\bibitem[\protect\citeauthoryear{Chen, Conroy, and Rubin}{Chen
  et~al\mbox{.}}{2015}]%
        {chen2015misleading}
\bibfield{author}{\bibinfo{person}{Yimin Chen}, \bibinfo{person}{Niall~J
  Conroy}, {and} \bibinfo{person}{Victoria~L Rubin}.}
  \bibinfo{year}{2015}\natexlab{}.
\newblock \showarticletitle{Misleading online content: Recognizing clickbait as
  false news}. In \bibinfo{booktitle}{{\em Proceedings of the 2015 ACM on
  Workshop on Multimodal Deception Detection}}. ACM, \bibinfo{pages}{15--19}.
\newblock


\bibitem[\protect\citeauthoryear{Edkins}{Edkins}{2016}]%
        {forbes}
\bibfield{author}{\bibinfo{person}{Brett Edkins}.}
  \bibinfo{year}{2016}\natexlab{}.
\newblock \bibinfo{title}{Americans Believe They Can Detect Fake News. Studies
  Show They Can't.}
\newblock   (\bibinfo{date}{December} \bibinfo{year}{2016}).
\newblock
\showURL{%
\url{www.forbes.com/sites/brettedkins/2016/12/20/americans-believe-they-can-detect-fake-news-studies-show-they-cant/}}


\bibitem[\protect\citeauthoryear{Feng and Hirst}{Feng and Hirst}{2013}]%
        {feng2013detecting}
\bibfield{author}{\bibinfo{person}{Vanessa~Wei Feng} {and}
  \bibinfo{person}{Graeme Hirst}.} \bibinfo{year}{2013}\natexlab{}.
\newblock \showarticletitle{Detecting Deceptive Opinions with Profile
  Compatibility.}. In \bibinfo{booktitle}{{\em IJCNLP}}.
  \bibinfo{pages}{338--346}.
\newblock


\bibitem[\protect\citeauthoryear{Ferreira and Vlachos}{Ferreira and
  Vlachos}{2016}]%
        {ferreira2016emergent}
\bibfield{author}{\bibinfo{person}{William Ferreira} {and}
  \bibinfo{person}{Andreas Vlachos}.} \bibinfo{year}{2016}\natexlab{}.
\newblock \showarticletitle{Emergent: a novel data-set for stance
  classification}. In \bibinfo{booktitle}{{\em Proceedings of the 2016
  Conference of the North American Chapter of the Association for Computational
  Linguistics: Human Language Technologies}}. ACL.
\newblock


\bibitem[\protect\citeauthoryear{Friggeri, Adamic, Eckles, and Cheng}{Friggeri
  et~al\mbox{.}}{2014}]%
        {friggeri2014rumor}
\bibfield{author}{\bibinfo{person}{Adrien Friggeri}, \bibinfo{person}{Lada~A
  Adamic}, \bibinfo{person}{Dean Eckles}, {and} \bibinfo{person}{Justin
  Cheng}.} \bibinfo{year}{2014}\natexlab{}.
\newblock \showarticletitle{Rumor Cascades.}. In \bibinfo{booktitle}{{\em
  ICWSM}}.
\newblock


\bibitem[\protect\citeauthoryear{Gupta, Kumaraguru, Castillo, and Meier}{Gupta
  et~al\mbox{.}}{2014}]%
        {gupta2014tweetcred}
\bibfield{author}{\bibinfo{person}{Aditi Gupta}, \bibinfo{person}{Ponnurangam
  Kumaraguru}, \bibinfo{person}{Carlos Castillo}, {and}
  \bibinfo{person}{Patrick Meier}.} \bibinfo{year}{2014}\natexlab{}.
\newblock \showarticletitle{Tweetcred: Real-time credibility assessment of
  content on twitter}. In \bibinfo{booktitle}{{\em International Conference on
  Social Informatics}}. Springer, \bibinfo{pages}{228--243}.
\newblock


\bibitem[\protect\citeauthoryear{H{\"u}sken and Stagge}{H{\"u}sken and
  Stagge}{2003}]%
        {husken2003recurrent}
\bibfield{author}{\bibinfo{person}{Michael H{\"u}sken} {and}
  \bibinfo{person}{Peter Stagge}.} \bibinfo{year}{2003}\natexlab{}.
\newblock \showarticletitle{Recurrent neural networks for time series
  classification}.
\newblock \bibinfo{journal}{{\em Neurocomputing\/}}  \bibinfo{volume}{50}
  (\bibinfo{year}{2003}), \bibinfo{pages}{223--235}.
\newblock


\bibitem[\protect\citeauthoryear{Jiang, Cui, and Faloutsos}{Jiang
  et~al\mbox{.}}{2016}]%
        {jiang2016suspicious}
\bibfield{author}{\bibinfo{person}{Meng Jiang}, \bibinfo{person}{Peng Cui},
  {and} \bibinfo{person}{Christos Faloutsos}.} \bibinfo{year}{2016}\natexlab{}.
\newblock \showarticletitle{Suspicious behavior detection: Current trends and
  future directions}.
\newblock \bibinfo{journal}{{\em IEEE Intelligent Systems\/}}
  \bibinfo{volume}{31}, \bibinfo{number}{1} (\bibinfo{year}{2016}),
  \bibinfo{pages}{31--39}.
\newblock


\bibitem[\protect\citeauthoryear{Jin, Dougherty, Saraf, Cao, and
  Ramakrishnan}{Jin et~al\mbox{.}}{2013}]%
        {jin2013epidemiological}
\bibfield{author}{\bibinfo{person}{Fang Jin}, \bibinfo{person}{Edward
  Dougherty}, \bibinfo{person}{Parang Saraf}, \bibinfo{person}{Yang Cao}, {and}
  \bibinfo{person}{Naren Ramakrishnan}.} \bibinfo{year}{2013}\natexlab{}.
\newblock \showarticletitle{Epidemiological modeling of news and rumors on
  twitter}. In \bibinfo{booktitle}{{\em Proceedings of the 7th Workshop on
  Social Network Mining and Analysis}}. ACM, \bibinfo{pages}{8}.
\newblock


\bibitem[\protect\citeauthoryear{Kumar, West, and Leskovec}{Kumar
  et~al\mbox{.}}{2016}]%
        {kumar2016disinformation}
\bibfield{author}{\bibinfo{person}{Srijan Kumar}, \bibinfo{person}{Robert
  West}, {and} \bibinfo{person}{Jure Leskovec}.}
  \bibinfo{year}{2016}\natexlab{}.
\newblock \showarticletitle{Disinformation on the web: Impact, characteristics,
  and detection of wikipedia hoaxes}. In \bibinfo{booktitle}{{\em Proceedings
  of the 25th International Conference on World Wide Web}}. International World
  Wide Web Conferences Steering Committee, \bibinfo{pages}{591--602}.
\newblock


\bibitem[\protect\citeauthoryear{Kwon, Cha, and Jung}{Kwon
  et~al\mbox{.}}{2017}]%
        {kwon2017rumor}
\bibfield{author}{\bibinfo{person}{Sejeong Kwon}, \bibinfo{person}{Meeyoung
  Cha}, {and} \bibinfo{person}{Kyomin Jung}.} \bibinfo{year}{2017}\natexlab{}.
\newblock \showarticletitle{Rumor Detection over Varying Time Windows}.
\newblock \bibinfo{journal}{{\em PLOS ONE\/}} \bibinfo{volume}{12},
  \bibinfo{number}{1} (\bibinfo{year}{2017}), \bibinfo{pages}{e0168344}.
\newblock


\bibitem[\protect\citeauthoryear{Le and Mikolov}{Le and Mikolov}{2014}]%
        {le2014distributed}
\bibfield{author}{\bibinfo{person}{Quoc~V Le} {and} \bibinfo{person}{Tomas
  Mikolov}.} \bibinfo{year}{2014}\natexlab{}.
\newblock \showarticletitle{Distributed Representations of Sentences and
  Documents.}. In \bibinfo{booktitle}{{\em ICML}}, Vol.~\bibinfo{volume}{14}.
  \bibinfo{pages}{1188--1196}.
\newblock


\bibitem[\protect\citeauthoryear{Lee and Dernoncourt}{Lee and
  Dernoncourt}{2016}]%
        {lee2016sequential}
\bibfield{author}{\bibinfo{person}{Ji~Young Lee} {and} \bibinfo{person}{Franck
  Dernoncourt}.} \bibinfo{year}{2016}\natexlab{}.
\newblock \showarticletitle{Sequential short-text classification with recurrent
  and convolutional neural networks}.
\newblock \bibinfo{journal}{{\em arXiv preprint arXiv:1603.03827\/}}
  (\bibinfo{year}{2016}).
\newblock


\bibitem[\protect\citeauthoryear{Leskovec, Rajaraman, and Ullman}{Leskovec
  et~al\mbox{.}}{2014}]%
        {leskovec2014mining}
\bibfield{author}{\bibinfo{person}{Jure Leskovec}, \bibinfo{person}{Anand
  Rajaraman}, {and} \bibinfo{person}{Jeffrey~David Ullman}.}
  \bibinfo{year}{2014}\natexlab{}.
\newblock \bibinfo{booktitle}{{\em Mining of massive datasets}}.
\newblock \bibinfo{publisher}{Cambridge University Press}.
\newblock


\bibitem[\protect\citeauthoryear{Lotan}{Lotan}{2016}]%
        {points}
\bibfield{author}{\bibinfo{person}{Gilad Lotan}.}
  \bibinfo{year}{2016}\natexlab{}.
\newblock \bibinfo{title}{Fake News Is Not the Only Problem.}
\newblock   (\bibinfo{date}{November} \bibinfo{year}{2016}).
\newblock
\showURL{%
\url{points.datasociety.net/fake-news-is-not-the-problem-f00ec8cdfcb}}


\bibitem[\protect\citeauthoryear{Luo, Tay, and Leng}{Luo et~al\mbox{.}}{2013}]%
        {luo2013identifying}
\bibfield{author}{\bibinfo{person}{Wuqiong Luo}, \bibinfo{person}{Wee~Peng
  Tay}, {and} \bibinfo{person}{Mei Leng}.} \bibinfo{year}{2013}\natexlab{}.
\newblock \showarticletitle{Identifying infection sources and regions in large
  networks}.
\newblock \bibinfo{journal}{{\em IEEE Transactions on Signal Processing\/}}
  \bibinfo{volume}{61}, \bibinfo{number}{11} (\bibinfo{year}{2013}),
  \bibinfo{pages}{2850--2865}.
\newblock


\bibitem[\protect\citeauthoryear{Ma, Gao, Mitra, Kwon, Jansen, Wong, and
  Cha}{Ma et~al\mbox{.}}{2016}]%
        {ma2016detecting}
\bibfield{author}{\bibinfo{person}{Jing Ma}, \bibinfo{person}{Wei Gao},
  \bibinfo{person}{Prasenjit Mitra}, \bibinfo{person}{Sejeong Kwon},
  \bibinfo{person}{Bernard~J Jansen}, \bibinfo{person}{Kam-Fai Wong}, {and}
  \bibinfo{person}{Meeyoung Cha}.} \bibinfo{year}{2016}\natexlab{}.
\newblock \showarticletitle{Detecting rumors from microblogs with recurrent
  neural networks}. In \bibinfo{booktitle}{{\em Proceedings of IJCAI}}.
\newblock


\bibitem[\protect\citeauthoryear{Ma, Gao, Wei, Lu, and Wong}{Ma
  et~al\mbox{.}}{2015}]%
        {ma2015detect}
\bibfield{author}{\bibinfo{person}{Jing Ma}, \bibinfo{person}{Wei Gao},
  \bibinfo{person}{Zhongyu Wei}, \bibinfo{person}{Yueming Lu}, {and}
  \bibinfo{person}{Kam-Fai Wong}.} \bibinfo{year}{2015}\natexlab{}.
\newblock \showarticletitle{Detect rumors using time series of social context
  information on microblogging websites}. In \bibinfo{booktitle}{{\em
  Proceedings of the 24th ACM International on Conference on Information and
  Knowledge Management}}. ACM, \bibinfo{pages}{1751--1754}.
\newblock


\bibitem[\protect\citeauthoryear{Maheshwari}{Maheshwari}{2016}]%
        {nyt}
\bibfield{author}{\bibinfo{person}{Sapa Maheshwari}.}
  \bibinfo{year}{2016}\natexlab{}.
\newblock \bibinfo{title}{How Fake News Goes Viral: A Case Study.}
\newblock   (\bibinfo{date}{November} \bibinfo{year}{2016}).
\newblock
\showURL{%
\url{https://www.nytimes.com/2016/11/20/business/media/how-fake-news-spreads.html}}


\bibitem[\protect\citeauthoryear{Markines, Cattuto, and Menczer}{Markines
  et~al\mbox{.}}{2009}]%
        {markines2009social}
\bibfield{author}{\bibinfo{person}{Benjamin Markines}, \bibinfo{person}{Ciro
  Cattuto}, {and} \bibinfo{person}{Filippo Menczer}.}
  \bibinfo{year}{2009}\natexlab{}.
\newblock \showarticletitle{Social spam detection}. In \bibinfo{booktitle}{{\em
  Proceedings of the 5th International Workshop on Adversarial Information
  Retrieval on the Web}}. ACM, \bibinfo{pages}{41--48}.
\newblock


\bibitem[\protect\citeauthoryear{Markowitz and Hancock}{Markowitz and
  Hancock}{2014}]%
        {markowitz2014linguistic}
\bibfield{author}{\bibinfo{person}{David~M Markowitz} {and}
  \bibinfo{person}{Jeffrey~T Hancock}.} \bibinfo{year}{2014}\natexlab{}.
\newblock \showarticletitle{Linguistic traces of a scientific fraud: The case
  of Diederik Stapel}.
\newblock \bibinfo{journal}{{\em PloS one\/}} \bibinfo{volume}{9},
  \bibinfo{number}{8} (\bibinfo{year}{2014}), \bibinfo{pages}{e105937}.
\newblock


\bibitem[\protect\citeauthoryear{McClure}{McClure}{2017}]%
        {ted}
\bibfield{author}{\bibinfo{person}{Laura McClure}.}
  \bibinfo{year}{2017}\natexlab{}.
\newblock \bibinfo{title}{How to tell fake news from real news.}
\newblock   (\bibinfo{date}{January} \bibinfo{year}{2017}).
\newblock
\showURL{%
\url{blog.ed.ted.com/2017/01/12/how-to-tell-fake-news-from-real-news/}}


\bibitem[\protect\citeauthoryear{Muandet and Sch{\"o}lkopf}{Muandet and
  Sch{\"o}lkopf}{2013}]%
        {muandet2013one}
\bibfield{author}{\bibinfo{person}{Krikamol Muandet} {and}
  \bibinfo{person}{Bernhard Sch{\"o}lkopf}.} \bibinfo{year}{2013}\natexlab{}.
\newblock \showarticletitle{One-class support measure machines for group
  anomaly detection}.
\newblock \bibinfo{journal}{{\em arXiv preprint arXiv:1303.0309\/}}
  (\bibinfo{year}{2013}).
\newblock


\bibitem[\protect\citeauthoryear{Mukherjee, Liu, and Glance}{Mukherjee
  et~al\mbox{.}}{2012}]%
        {mukherjee2012spotting}
\bibfield{author}{\bibinfo{person}{Arjun Mukherjee}, \bibinfo{person}{Bing
  Liu}, {and} \bibinfo{person}{Natalie Glance}.}
  \bibinfo{year}{2012}\natexlab{}.
\newblock \showarticletitle{Spotting fake reviewer groups in consumer reviews}.
  In \bibinfo{booktitle}{{\em Proceedings of the 21st international conference
  on World Wide Web}}. ACM, \bibinfo{pages}{191--200}.
\newblock


\bibitem[\protect\citeauthoryear{Preli{\'c}, Bleuler, Zimmermann, Wille,
  B{\"u}hlmann, Gruissem, Hennig, Thiele, and Zitzler}{Preli{\'c}
  et~al\mbox{.}}{2006}]%
        {prelic2006systematic}
\bibfield{author}{\bibinfo{person}{Amela Preli{\'c}}, \bibinfo{person}{Stefan
  Bleuler}, \bibinfo{person}{Philip Zimmermann}, \bibinfo{person}{Anja Wille},
  \bibinfo{person}{Peter B{\"u}hlmann}, \bibinfo{person}{Wilhelm Gruissem},
  \bibinfo{person}{Lars Hennig}, \bibinfo{person}{Lothar Thiele}, {and}
  \bibinfo{person}{Eckart Zitzler}.} \bibinfo{year}{2006}\natexlab{}.
\newblock \showarticletitle{A systematic comparison and evaluation of
  biclustering methods for gene expression data}.
\newblock \bibinfo{journal}{{\em Bioinformatics\/}} \bibinfo{volume}{22},
  \bibinfo{number}{9} (\bibinfo{year}{2006}), \bibinfo{pages}{1122--1129}.
\newblock


\bibitem[\protect\citeauthoryear{Rubin}{Rubin}{2017}]%
        {rubin2017deception}
\bibfield{author}{\bibinfo{person}{Victoria~L Rubin}.}
  \bibinfo{year}{2017}\natexlab{}.
\newblock \showarticletitle{Deception Detection and Rumor Debunking for Social
  Media}.
\newblock  (\bibinfo{year}{2017}).
\newblock


\bibitem[\protect\citeauthoryear{Rubin, Chen, and Conroy}{Rubin
  et~al\mbox{.}}{2015}]%
        {rubin2015deception}
\bibfield{author}{\bibinfo{person}{Victoria~L Rubin}, \bibinfo{person}{Yimin
  Chen}, {and} \bibinfo{person}{Niall~J Conroy}.}
  \bibinfo{year}{2015}\natexlab{}.
\newblock \showarticletitle{Deception detection for news: three types of
  fakes}.
\newblock \bibinfo{journal}{{\em Proceedings of the Association for Information
  Science and Technology\/}} \bibinfo{volume}{52}, \bibinfo{number}{1}
  (\bibinfo{year}{2015}), \bibinfo{pages}{1--4}.
\newblock


\bibitem[\protect\citeauthoryear{Starbird, Maddock, Orand, Achterman, and
  Mason}{Starbird et~al\mbox{.}}{2014}]%
        {starbird2014rumors}
\bibfield{author}{\bibinfo{person}{Kate Starbird}, \bibinfo{person}{Jim
  Maddock}, \bibinfo{person}{Mania Orand}, \bibinfo{person}{Peg Achterman},
  {and} \bibinfo{person}{Robert~M Mason}.} \bibinfo{year}{2014}\natexlab{}.
\newblock \showarticletitle{Rumors, false flags, and digital vigilantes:
  Misinformation on twitter after the 2013 boston marathon bombing}.
\newblock \bibinfo{journal}{{\em iConference 2014 Proceedings\/}}
  (\bibinfo{year}{2014}).
\newblock


\bibitem[\protect\citeauthoryear{Sutskever, Vinyals, and Le}{Sutskever
  et~al\mbox{.}}{2014}]%
        {sutskever2014sequence}
\bibfield{author}{\bibinfo{person}{Ilya Sutskever}, \bibinfo{person}{Oriol
  Vinyals}, {and} \bibinfo{person}{Quoc~V Le}.}
  \bibinfo{year}{2014}\natexlab{}.
\newblock \showarticletitle{Sequence to sequence learning with neural
  networks}. In \bibinfo{booktitle}{{\em Advances in neural information
  processing systems}}. \bibinfo{pages}{3104--3112}.
\newblock


\bibitem[\protect\citeauthoryear{Townsend}{Townsend}{2017}]%
        {google}
\bibfield{author}{\bibinfo{person}{Tess Townsend}.}
  \bibinfo{year}{2017}\natexlab{}.
\newblock \bibinfo{title}{Google has banned 200 publishers since it passed a
  new policy against fake news.}
\newblock   (\bibinfo{date}{January} \bibinfo{year}{2017}).
\newblock
\showURL{%
\url{www.recode.net/2017/1/25/14375750/google-adsense-advertisers-publishers-fake-news}}


\bibitem[\protect\citeauthoryear{Varol, Ferrara, Davis, Menczer, and
  Flammini}{Varol et~al\mbox{.}}{2017}]%
        {varol2017online}
\bibfield{author}{\bibinfo{person}{Onur Varol}, \bibinfo{person}{Emilio
  Ferrara}, \bibinfo{person}{Clayton~A Davis}, \bibinfo{person}{Filippo
  Menczer}, {and} \bibinfo{person}{Alessandro Flammini}.}
  \bibinfo{year}{2017}\natexlab{}.
\newblock \showarticletitle{Online human-bot interactions: Detection,
  estimation, and characterization}.
\newblock \bibinfo{journal}{{\em arXiv preprint arXiv:1703.03107\/}}
  (\bibinfo{year}{2017}).
\newblock


\bibitem[\protect\citeauthoryear{Wang and Nyberg}{Wang and Nyberg}{2015}]%
        {wang2015long}
\bibfield{author}{\bibinfo{person}{Di Wang} {and} \bibinfo{person}{Eric
  Nyberg}.} \bibinfo{year}{2015}\natexlab{}.
\newblock \showarticletitle{A Long Short-Term Memory Model for Answer Sentence
  Selection in Question Answering.}. In \bibinfo{booktitle}{{\em ACL (2)}}.
  \bibinfo{pages}{707--712}.
\newblock


\bibitem[\protect\citeauthoryear{Wang, Dong, Zhang, and Tan}{Wang
  et~al\mbox{.}}{2014}]%
        {wang2014rumor}
\bibfield{author}{\bibinfo{person}{Zhaoxu Wang}, \bibinfo{person}{Wenxiang
  Dong}, \bibinfo{person}{Wenyi Zhang}, {and} \bibinfo{person}{Chee~Wei Tan}.}
  \bibinfo{year}{2014}\natexlab{}.
\newblock \showarticletitle{Rumor source detection with multiple observations:
  Fundamental limits and algorithms}. In \bibinfo{booktitle}{{\em ACM
  SIGMETRICS Performance Evaluation Review}}, Vol.~\bibinfo{volume}{42}. ACM,
  \bibinfo{pages}{1--13}.
\newblock


\bibitem[\protect\citeauthoryear{Wu, Yang, and Zhu}{Wu et~al\mbox{.}}{2015}]%
        {wu2015false}
\bibfield{author}{\bibinfo{person}{Ke Wu}, \bibinfo{person}{Song Yang}, {and}
  \bibinfo{person}{Kenny~Q Zhu}.} \bibinfo{year}{2015}\natexlab{}.
\newblock \showarticletitle{False rumors detection on sina weibo by propagation
  structures}. In \bibinfo{booktitle}{{\em Data Engineering (ICDE), 2015 IEEE
  31st International Conference on}}. IEEE, \bibinfo{pages}{651--662}.
\newblock


\bibitem[\protect\citeauthoryear{Xiong, P{\'o}czos, and Schneider}{Xiong
  et~al\mbox{.}}{2011a}]%
        {xiong2011group}
\bibfield{author}{\bibinfo{person}{Liang Xiong}, \bibinfo{person}{Barnab{\'a}s
  P{\'o}czos}, {and} \bibinfo{person}{Jeff~G Schneider}.}
  \bibinfo{year}{2011}\natexlab{a}.
\newblock \showarticletitle{Group anomaly detection using flexible genre
  models}. In \bibinfo{booktitle}{{\em Advances in neural information
  processing systems}}. \bibinfo{pages}{1071--1079}.
\newblock


\bibitem[\protect\citeauthoryear{Xiong, P{\'o}czos, Schneider, Connolly, and
  VanderPlas}{Xiong et~al\mbox{.}}{2011b}]%
        {xiong2011hierarchical}
\bibfield{author}{\bibinfo{person}{Liang Xiong}, \bibinfo{person}{Barnab{\'a}s
  P{\'o}czos}, \bibinfo{person}{Jeff~G Schneider}, \bibinfo{person}{Andrew~J
  Connolly}, {and} \bibinfo{person}{Jake VanderPlas}.}
  \bibinfo{year}{2011}\natexlab{b}.
\newblock \showarticletitle{Hierarchical Probabilistic Models for Group Anomaly
  Detection.}. In \bibinfo{booktitle}{{\em AISTATS}}.
  \bibinfo{pages}{789--797}.
\newblock


\bibitem[\protect\citeauthoryear{Yu, He, and Liu}{Yu et~al\mbox{.}}{2015}]%
        {yu2015glad}
\bibfield{author}{\bibinfo{person}{Rose Yu}, \bibinfo{person}{Xinran He}, {and}
  \bibinfo{person}{Yan Liu}.} \bibinfo{year}{2015}\natexlab{}.
\newblock \showarticletitle{Glad: group anomaly detection in social media
  analysis}.
\newblock \bibinfo{journal}{{\em ACM Transactions on Knowledge Discovery from
  Data (TKDD)\/}} \bibinfo{volume}{10}, \bibinfo{number}{2}
  (\bibinfo{year}{2015}), \bibinfo{pages}{18}.
\newblock


\bibitem[\protect\citeauthoryear{Zhao, Resnick, and Mei}{Zhao
  et~al\mbox{.}}{2015}]%
        {zhao2015enquiring}
\bibfield{author}{\bibinfo{person}{Zhe Zhao}, \bibinfo{person}{Paul Resnick},
  {and} \bibinfo{person}{Qiaozhu Mei}.} \bibinfo{year}{2015}\natexlab{}.
\newblock \showarticletitle{Enquiring minds: Early detection of rumors in
  social media from enquiry posts}. In \bibinfo{booktitle}{{\em Proceedings of
  the 24th International Conference on World Wide Web}}. ACM,
  \bibinfo{pages}{1395--1405}.
\newblock


\bibitem[\protect\citeauthoryear{Zhu and Ying}{Zhu and Ying}{2016}]%
        {zhu2016information}
\bibfield{author}{\bibinfo{person}{Kai Zhu} {and} \bibinfo{person}{Lei Ying}.}
  \bibinfo{year}{2016}\natexlab{}.
\newblock \showarticletitle{Information source detection in the SIR model: A
  sample-path-based approach}.
\newblock \bibinfo{journal}{{\em IEEE/ACM Transactions on Networking (TON)\/}}
  \bibinfo{volume}{24}, \bibinfo{number}{1} (\bibinfo{year}{2016}),
  \bibinfo{pages}{408--421}.
\newblock


\end{thebibliography}
